# The Role of LLMs in Sustainable Smart Cities: Applications, Challenges, and Future Directions

Amin Ullah[1], Guilin Qi[1], Saddam Hussain[2], Irfan Ullah[3], Zafar Ali[1]

[1]School of Computer Science and Engineering, Southeast University (SEU), Nanjing, China. aminullah@seu.edu.cn

[2]Artificial Intelligence Lab, Department of Computer Systems Engineering, University of Engineering and Applied Sciences (UEAS), Swat, Pakistan.

[3]Department of Computer Science, Shaheed Benazir Bhutto University, Sheringal, Upper Dir, Pakistan.

## ABSTRACT

Smart cities stand as pivotal components in the ongoing pursuit of elevating urban living standards, facilitating the rapid expansion of urban areas while efficiently managing resources through sustainable and scalable innovations. In this regard, as emerging technologies like Artificial Intelligence (AI), the Internet of Things (IoT), big data analytics, and fog and edge computing have become increasingly prevalent, smart city applications grapple with various challenges, including the potential for unauthorized disclosure of confidential and sensitive data. The seamless integration of emerging technologies has played a vital role in sustaining the dynamic pace of their development. This paper explores the substantial potential and applications of Deep Learning (DL), Federated Learning (FL), IoT, Blockchain, Natural Language Processing (NLP), and large language models (LLMs) in optimizing ICT processes within smart cities. We aim to spotlight the vast potential of these technologies as foundational elements that technically strengthen the realization and advancement of smart cities, underscoring their significance in driving innovation within this transformative urban milieu. Our discourse culminates with an exploration of the formidable challenges that DL, FL, IoT, Blockchain, NLP, and LLMs face within these contexts, and we offer insights into potential future directions.

**Keywords**: Sustainable Smart Cities, Deep Learning, Federated Learning, Blockchain, Natural Language Processing, Large Language Models

## 1 INTRODUCTION

Smart cities have emerged as the vanguard of urban development, arranging efficient urbanization, environmental stewardship, enhanced living standards, and harnessing the transformative power of cutting-edge communication technologies [1]. The magnetic pull of urban opportunities has seen rural populations gravitate towards city centers, a trend projected to continue. The United Nations estimates that by 2050, 86% of developed and 64% of developing countries will have urbanized [2]. Similarly, by mid-century, globally, about 66% of people are projected to live in urban landscapes [3].

To tackle the rising challenges that accompany urban growth, policymakers in smart cities are compelled to craft innovative and well-calibrated solutions. The burgeoning Information and Communication Technology (ICT) sector stands as a pillar fortifying the sustainable future of smart cities. The transformative potential of ICT spans the spectrum, from advancing healthcare systems, optimizing water and energy and management, revolutionizing public transportation, and elevating tourism experiences to streamlining food supply chains and delivering high-caliber education [4].

As depicted in Figure 1, ICT has a paramount importance in the smart city narrative [5]. This recognition hinges on the fundamental premise that smart cities depend on the ICT infrastructure for retrieving, processing, and exchanging data. Nevertheless, the escalating reliance on ICT and IoT devices ushers in an array of challenges. IoT devices, functioning autonomously, gather, process, and exchange data with their peers, rendering data susceptible to nefarious cyber [6][7]. Additionally, IoT devices grapple with issues such as false data injection attacks and vulnerability to single-point failures during cloud-based data exchanges [8]–[10]. Moreover, the sole reliance on cloud-based security measures may prove inadequate to safeguard complicated and complex smart city applications.

Enterprising technologies and tools, including Artificial Intelligence (AI), ML, DL, IoT, Federated Learning (FL), Blockchain, NLP, and LLMs stand as indispensable sentinels, ensuring the seamless and standardized operation of smart cities [1]. In the realm of smart cities, AI leverages advanced algorithms to enable machines to mimic human intelligence, allowing them to learn from data, identify patterns, and make decisions [11]. Machine learning empowers systems to learn and improve from experience without explicit programming, enabling them to automatically identify hidden patterns and make data-driven predictions [12]. Deep Learning, under the umbrella of ML, models complex patterns and representation by employing neural networks with multiple layers in processes like recognizing speech and images [13]. Federated Learning is a collaborative ML approach that trains its models on decentralized data sources without sharing raw data, ensuring privacy while building global models [14]. Natural Language Processing involves AI algorithms in understanding, interpreting, and generating human language, facilitating interactions between humans and machine [15]. Blockchain technology is a decentralized and secure digital ledger that records transactions across multiple computers, enhancing data transparency and security [16].

Empowered by these technologies, administrators and city planners can orchestrate timely and precise interventions to enhance the overall quality of life for their residents [12],[13]. This synergy between cutting-edge technologies and urban governance paves the way for more efficient, responsive, and citizen-centric urban ecosystems, where AI, ML, DL, FL, NLP, LLMS, and Blockchain collectively play pivotal roles in shaping the smart cities of the future, as illustrated in Figure 1. Within the domain of smart cities, numerous technological advancements bring forth unprecedented opportunities and challenges. Addressing these challenges requires innovative solutions, such as FL and advanced security measures.

## 1.1 PROBLEM STATEMENT

In the dynamic landscape of smart city development, the integration of AI, DL, FL, IoT, Blockchain, and LLMs presents a promising avenue for sustainable urban growth, as presented in Figure 1. However, the journey towards these sustainable urban paradigms is riddled with complexities [19]. Challenges encompass not only the development of robust AI models but also the preservation of privacy and security in the era of FL and Blockchain. Furthermore, the integration of IoT devices demands careful orchestration to ensure seamless data flow and interpretation. Additionally, LLMs, while promising for natural language understanding and interaction, pose challenges related to bias, ethics, and effective utilization [20]. The complex nature of these technologies necessitates a complete survey that identifies key applications, analyzes pressing challenges, and outlines future directions. This survey paper aims to provide a comprehensive overview of the opportunities and obstacles in deploying DL, FL, IoT, Blockchain, and LLMs for sustainable smart cities, ultimately guiding researchers and practitioners toward innovative and informed solutions in this transformative domain. It addresses the need to harness the power of AI and DL to optimize city operations, ensure resource efficiency, and enhance the quality of life for urban inhabitants.

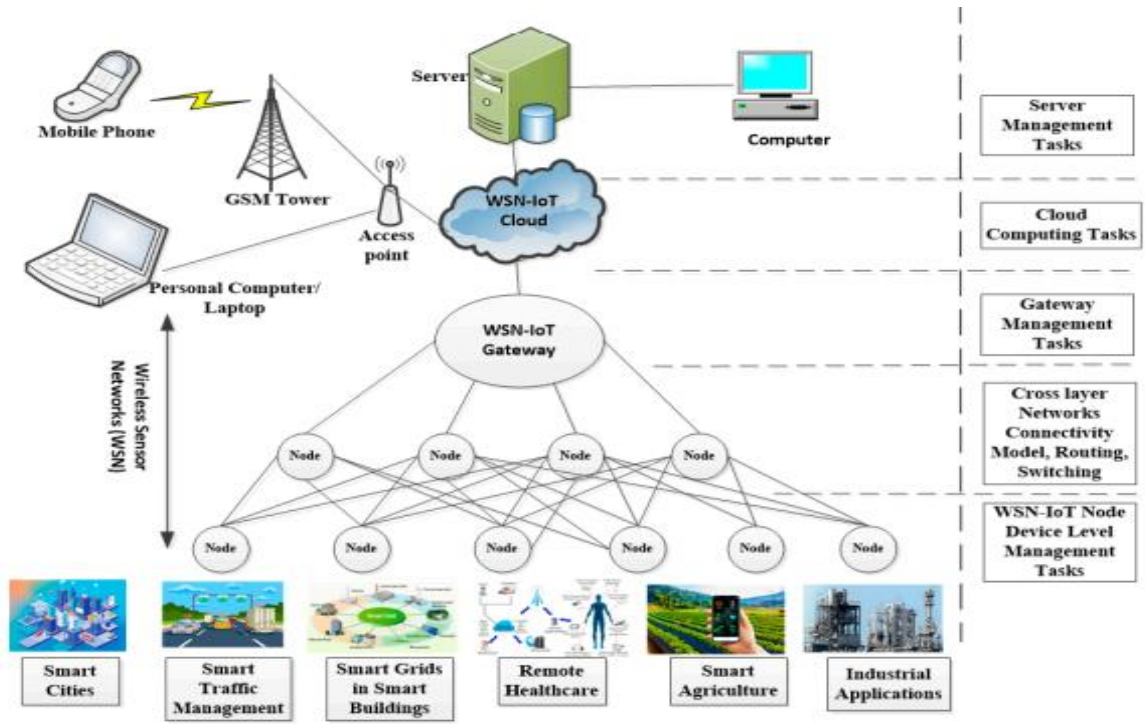

**Figure 1.** The dynamic landscape of sustainable smart city development by integrating AI, DL, FL, IoT, Blockchain, and LLMs [19]

## 1.2 CONTRIBUTIONS

The quest for enhancing the quality of life, efficiency, and sustainability in urban settings is of paramount importance. It prompts us to explore the literature on smart cities. This survey highlights the implications of technological advancements for sustainable smart cities. Specifically, the following are the key contributions of this survey.

- We present an overview of the research developments in DL, FL, Blockchain, IoT, and LLMs for sustainable smart cities.
- We highlight how emerging technologies can be integrated to sustain the dynamic pace of smart city developments by exploring the optimal solutions for intelligent resource allocation that consider energy efficiency, packet reliability, and latency, all while weighing their trade-offs.
- We report on various applications of these algorithms, including, e.g., smart transportation, energy management, healthcare, security, public safety, and environmental sustainability. The findings have implications for specific domains such as smart grids, healthcare systems, autonomous driving, crime control, and other distributed systems.

- We identify and report on the current challenges, limitations, and barriers associated with adopting and implementing these technologies in smart cities. These include technical, regulatory, ethical, and social challenges. We highlight the potential for innovation and growth in sustainable smart cities.

## 1.3 ORGANIZATION OF THE PAPER

The rest of the paper spans ten sections. Section 2 presents the architecture, building blocks, and characteristics of smart cities. Section 3 presents the applications of DL in smart cities. Section 4 presents the role of Blockchain in sustainable smart cities. Section 5 presents the application of federated learning in sustainable smart cities. Section 6 presents the applications of NLP in sustainable smart cities. Section 7 presents LLMs and their applications in sustainable smart cities. Section 8 presents challenges in smart cities. Section 9 presents future directions. Section 10 concludes the paper followed by references.

## 2 SMART CITIES: ARCHITECTURE AND BUILDING BLOCKS

This section discusses the architecture and building blocks required for a sustainable smart city.

### 2.1 THE ARCHITECTURE OF A SMART CITY

Several studies have aimed to formulate a universal architecture for smart cities worldwide. Figure 2 shows a layered architecture of smart cities having layers namely data collection, data transmission, data management, and application layer [21]. This architecture effectively encapsulates the primary components that underpin smart city operations, allowing for comprehensive yet concise descriptions and planning of various smart city technologies.

#### 2.1.1 Data Collection Layer

This layer gathers data from diverse sources using physical tools such as cameras, microphones, GPS trackers, and other sensors. Collecting data from these sources is a complex task, as the acquired data is often unstructured and heterogeneous. Deploying and using wireless sensor networks (WSN) is among the key concerns across the smart city [22]. Other concerns regarding efficient data collection require the proper management of energy consumption, resource usage, and keeping transmission costs at a minimum. Delay tolerant networks (DTN), such as vehicular DTN, can enhance data collection in smart cities by employing methods to collect heterogeneous data from various applications [23].

### 2.1.2 Data Transmission Layer

The collected data is transmitted to storage units for further processing through suitable channels. Wireless networks, employing technologies like Wi-Fi, Bluetooth, Zigbee, and RFID, as well as telecommunication technologies like LTE, 3G, 4G, and 5G, are commonly employed for this purpose. For instance, in the context of smart healthcare, a recent work [24] presents an Interference Aware Energy Efficient Transmission Protocol (IEETP) that employs Wireless Body Area Networks (WBAN) to transmit data from sensors inside human bodies efficiently.

### 2.1.3 Data Management Layer

The received data is pre-processed, analyzed, and used in decision-making. Positioned between data acquisition and its application, this layer is often referred to as the "brain" of the smart city framework. Key tasks include filtering relevant data, combining it with data from diverse sources, and employing Big Data Analytics (BDA) for efficient real-time analysis. A recent study [18] applied BDA for sustainable development and making informed decisions in smart tourism. This layer uses ML and AI in decision-making.

### 2.1.4 Application Layer

This layer allows citizens to directly interact with the smart city ecosystem. Decisions made in the Data Management Layer are applied concurrently in this layer, where the primary aim is to create user-friendly and satisfactory applications. Communication plays a substantial role here. The applications span education, healthcare, management, transport, and related technical domains.

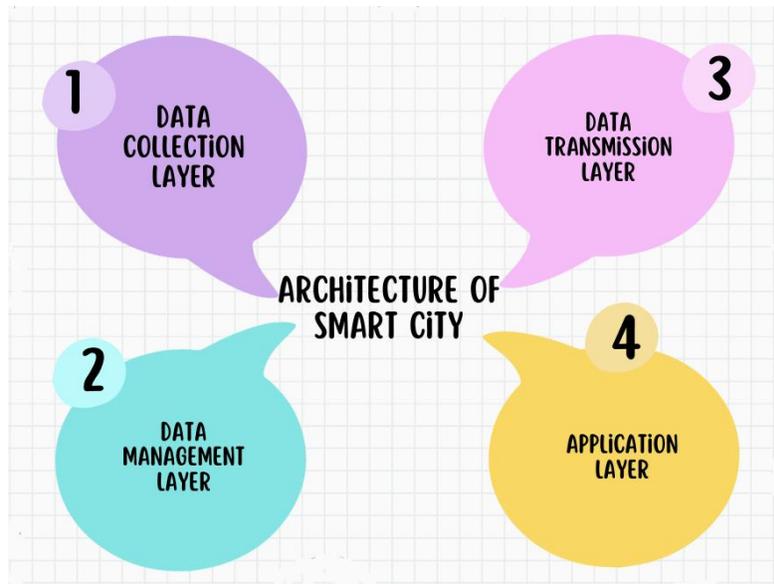

**Figure 2.** Architecture of Smart City

## 2.2 BUILDING BLOCKS OF SMART CITIES

A smart city's definition can be crafted through the consideration of various themes, characteristics, constituents, and requisites. Comprehending these factors not only facilitates the creation of more effective and sustainable strategies for these evolving urban areas but also simplifies the comparative evaluation of their smartness quotient. Within the context of strategic planning and governance that spans cities, regions, or even entire countries, the Political, Economic, Social, and Technological (PEST) analysis assumes a fundamental role as an initial step [25]. Similarly, the PEST analysis is also carried out for smart cities to empower them to adequately address present and future necessities across a wide array of sectors. A comprehensive model of the smart city encompasses various components and attributes, as depicted in Figure 3, including infrastructure, governance, energy, technology, education, buildings, healthcare, transportation, and community.

### 2.2.1 Infrastructure and Buildings

A smart city's infrastructure depends heavily on ICT and has domestic, public, and institutional segments. Domestic infrastructure focuses on households, incorporating smart appliances and resource management (water, electricity, Internet). Public infrastructure supports mobility and includes buildings, community centers, transportation, traffic management, and smart parking. Institutional infrastructure drives technological advancements, such as connectivity, data production and management, secure communication, and resource consumption [26].

### 2.2.2 Technology and Energy

Smart technology plays a crucial role in planning, implementing, and monitoring smart cities. Technological advancements and innovations help develop and manage infrastructure, resources, manufacturing, transportation, and connectivity. Transitioning from non-renewable to renewable energy sources like solar energy relies on technological advancements such as solar cells [27].

### 2.2.3 Transportation

Intelligent Transport Systems (ITS) are transforming conventional transportation through ICT, enhancing mobility and travel experiences in smart cities. The ITS principles emphasize sustainability, responsiveness, safety, and integration. Smart city transportation includes electronic and autonomous cars, advanced parking, and modern infrastructure that uses sensor processing and monitoring [28].

### 2.2.4 Education and Healthcare

Smart education integrates advanced pedagogical tools and ICT to offer effective learning experiences. The Internet of Things introduces devices such as smart boards, WSNs, and smartphones into educational institutions, alongside software services like learning management systems and enterprise resource planning. Smart healthcare encompasses services including intelligent tools and robots, emergency response systems, advanced diagnostics, and sophisticated hospital infrastructure. Analyzing medical records' data and enabling e-healthcare through smartphones are notable applications within smart city healthcare [22][23].

### 2.2.5 Community

The community comprises residents in a smart city. A community of smart citizens contributes to the advancement of efficient and sustainable modern cities. Smart city residents tend to seek a better quality of life compared to traditional city dwellers. This dynamic leads to a population of capable and knowledgeable individuals in smart cities, fostering further enhancement. These components collectively define the intricate framework of a smart city, encompassing diverse aspects that contribute to its growth and effectiveness.

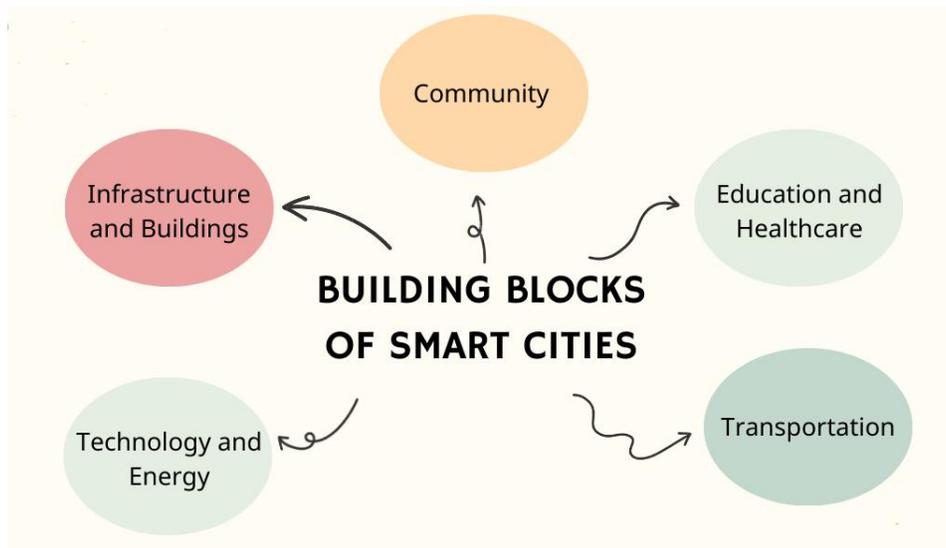

**Figure 3.** The building blocks of a sustainable smart city [31]

## 2.3 CHARACTERISTICS OF SMART CITIES

A smart city must ensure efficient and sustainable living space by designing its network-related services more flexibly and user-friendly [5]. This is where ICT comes in to improve residents' lifestyles and offer them a safe, green, and friendly environment so that their work efficiency gets improved [21].

Smart cities are characterized by multiple aspects and components, as shown in Figure 4. These include technology, livability, sustainability, governance, and urban aspects [21]. Among these, technology plays a vital role, where multiple components, including smart transportation, infrastructure, healthcare, technology, and energy play their role. The intelligence makes all these facilities efficient and fast while reducing cost and time waste. Information technology is the key to helping traditional cities transform into smart cities. The main technologies include but are not limited to IoT, big data, ICT, ML/DL algorithms, LLMs, NLP, and intelligent infrastructure.

The latest study [32] reports over two-thirds of the world population will be living in metropolitan regions by the end of the year 2050. The assets in urban communities are around 73% and they are additionally adding to the age of ozone-harming substances. This indicates a greater portion of the population by the consequences that can be influencing the climate, leading to the idea of smart urban areas [21]. The aim is to address the issues caused by massive expansion in urbanization, for example, bringing down energy utilization, water use, toxins emanation in climate, transportation use, and controlling city wastage. In addition, it is necessary to ensure personal satisfaction in terms of the prosperity of the monetary and passionate conditions of living spaces. In short, a smart city or smart urban area must have plans for climate, society, economy, and administration. It is because a smart city or urban area is intended for its residents, where the economy should make it possible to give them open positions, where the administration should ensure the necessary management actions for making smart cities or urban areas livable for their inhabitants through smart transportation, governance, energy, citizens, buildings, healthcare, and technology.

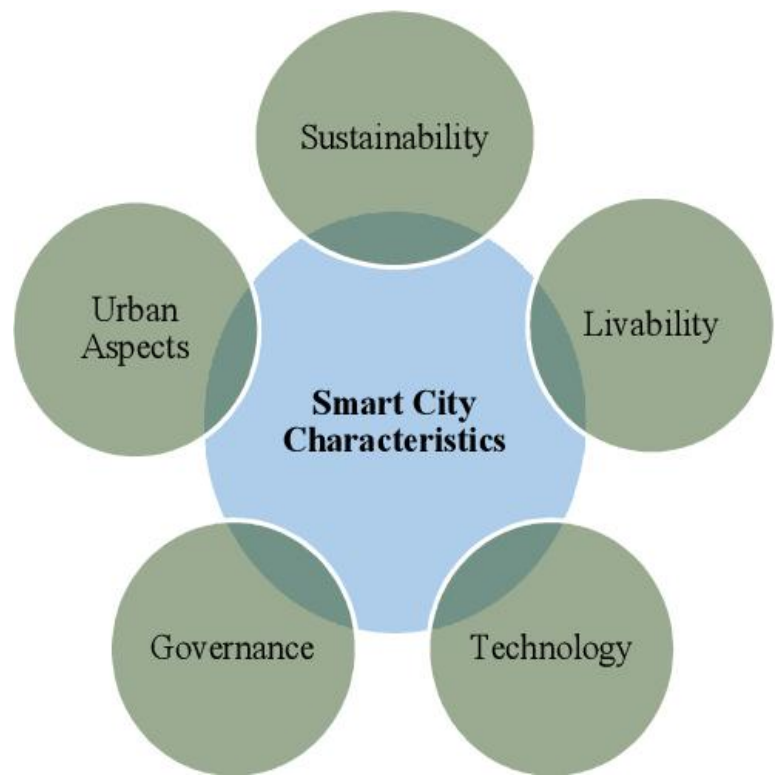

**Figure 4.** Smart city characteristics

## 3 DEEP LEARNING AND ITS APPLICATIONS FOR SUSTAINABLE SMART CITIES

Over the past five decades, the global urban population has witnessed a significant doubling, marking a notable demographic shift [33]. Urban and rural areas alike have been compelled to adapt to the rapid tide of technological innovation, unlocking fresh opportunities to enhance public safety, connectivity, and the overall experience for both residents and visitors.

In a groundbreaking study referenced in [34], researchers harnessed the power of sophisticated computer vision and DL patterns to detect and delineate shanty towns, as well as predict traffic rate expectations. This pioneering model aspires to identify a broad spectrum of gradations spanning from developed to underprivileged regions, contributing to a more nuanced understanding of urban landscapes. Moving forward to [35], a compelling recommendation emerges in the realm of smart parking, fortified by the intensive integration of DL, IoT, and wireless communication technologies. This visionary proposal sets its sights on drastically reducing the time it takes to locate available parking spaces, particularly in bustling locations like shopping malls and railway stations. By leveraging the capabilities of DL and IoT radar, this system swiftly pinpoints the optimal parking spots, thereby expediting the entire parking process. As urban landscapes continue to evolve, achieving seamless connectivity hinges on factors such as system availability, and network bandwidth, media, and topology [36].

Another ingenious innovation highlighted here is the implementation of smart road polls, bolstered by ML methods, which have the potential to yield up to 50% energy savings [31]. This sustainable approach to power distribution revolves around a holistic strategy encompassing power generation, distribution, and transmission. In essence, as urbanization progresses and technologies advance, the transformation of urban and rural spaces becomes increasingly intricate and interconnected. These endeavors underscore the profound impact that technology and innovation have on the urban landscape, with the potential to shape a more efficient, sustainable, and interconnected future.

## 3.1 INTELLIGENT INFRASTRUCTURE

The continuous urbanization trend is set to bring about a staggering reality by 2030, with 60% of the world's population dwelling in cities [37]. However, the rapid growth of urban centers has ushered in a slew of challenges, ranging from congestion and environmental unsustainability to a lack of resilience in the face of global warming. A beacon of hope emerges in the form of DL technologies that promise to reshape the very foundations of smart city infrastructures.

Pioneers in the field have harnessed DL prowess to craft an intelligent transportation model. This cutting-edge system assumes the role of a vigilant sentinel, monitoring energy consumption, and traffic flow, and making critical decisions based on the severity of the situation. The integration of DL empowers the infrastructure to proactively address issues and optimize city-wide transportation. An intelligent DL routing mechanism [38] is designed to tackle the deluge of data emanating from various sensors. This innovation acts as a remedy for network congestion, ensuring seamless data processing and analysis.

## 3.2 SMART MOBILITY AND TRANSPORTATION

Smart mobility and transportation are pivotal components of modern urban life, underpinned by a sophisticated network of cloud platforms and AI systems that interconnect vehicles, people, public services, and logistics partners. In this visionary landscape, autonomous vehicles equipped with radar systems continually monitor their surroundings, preemptively averting potential accidents and ensuring the safety of citizens. The convergence of DL, data analytics, and communication technologies delves into forging connections among people, roadways, and vehicles, addressing a myriad of traffic-related challenges. The focus here is on creating a vehicle-centric, safe, and comfortable transportation ecosystem. Safety takes center stage with innovations like a driver assistance system [21] designed to prevent red-light violations. Furthermore, the Pedestrian Detection System [39], fueled by AI and 3D stereo cameras, exhibits the remarkable ability to detect multiple pedestrians in its vicinity.

## 3.3 SMART TOWN GOVERNANCE

The urban shift of the populace has catalyzed the drive to build smarter cities, necessitating a comprehensive understanding of public management concepts. This understanding is instrumental in analyzing government policies and smart city features. Public opinion forms a critical element in shaping government strategies in the urban landscape.

## 3.4 STRENGTH AND SUSTAINABILITY

As information generation escalates, there's an imperative need for an efficient network to disseminate this knowledge, fostering the growth of smart cities. Environmental sustainability stands as a formidable challenge, compelling the creation of intelligent networks to reduce pollution levels and encourage healthier lifestyles [21]. An ingenious intelligent fault detection methodology employing clustering and DL techniques is used [40]. Additionally, a novel platform called SureCity24 [18] emerges on the horizon, aimed at addressing the existing challenges and facilitating the construction of smart, sustainable urban centers. This multifaceted approach is poised to revolutionize the way smart cities harness and share information while ensuring their long-term viability and vitality.

## 3.5 SMART EDUCATION

The landscape of education is undergoing a profound transformation with DL integration. In the realm of distance education for young adults, the online flipped classroom learning method [18], is ushering in new possibilities for remote learning scenarios. To gauge the effectiveness of educational content, more systems [18] were introduced for a face detection system that analyzes learners' facial emotions.

In the pursuit of enhancing learning experiences, an innovative emotion-sensitive approach [18] was developed. This method evaluates learners' interest by considering factors such as head position and facial expressions. A key challenge in online learning is retention, often influenced by content presentation and interaction limitations. A recent study [41] tackles a temporal sequential classification problem, predicting student retention rates by harnessing interactional events within the online learning system.

## 3.6 REVOLUTIONIZING HEALTHCARE

The marriage of AI and DL heralds a new era in healthcare solutions [42], [43]. Remarkable accuracy has been achieved in classifying and estimating breast cancer images using advanced DL techniques, surpassing traditional architectures. Similar advancements are observed in cervical cancer image segmentation [44], leveraging wireless network technology. Moreover, DL finds utility in MRI image segmentation, yielding results that nearly double the efficiency of conventional approaches. These developments contribute to automated systems that elevate the quality of life [45].

## 3.7 SECURITY AND PRIVACY

Smart cities, epitomes of technological innovation, offer unparalleled convenience and lifestyle enhancements through interconnected devices and IoT [21]. However, this interconnectedness brings forth security and privacy challenges, with big data and IoT at the forefront. Deep learning and related technologies rise to the occasion, providing solutions to mitigate security breaches. An example is the use of random forest algorithms in detecting anomalies in the IoT environment [46]. This technology protects distributed fog nodes within IoT devices so that the fog and cloud layers can communicate securely. Such solutions outperform traditional non-cooperative and semi-cooperative techniques, fortifying the smart city infrastructure.

Blockchain technology can potentially combat drug counterfeiting and revolutionize tracking and monitoring processes [47]. It can effectively manage demand response in the Internet of Vehicles [48], e-government, environments, supply chains, energy, and construction [49]. It facilitates new sustainable business models and aligns with the United Nations Sustainable Development Goals, especially in sustainable production, energy and waste management, and climate change [50]. Its potential spans energy management in smart grids, energy trading, peer-to-peer microgrids, decentralized energy management, energy market innovations, network data transmission, data security and cybersecurity, environmental concerns, IoT, testbeds, demand response, smart contracts [51][52][53], and Industry 4.0 [54]. Some applications and research domains are discussed in the following subsections.

### 4.1    FOOD INDUSTRY

Today's world faces an increasing demand for food supply and sustainability. This demand arises from a mix of factors, including the impacts of climate change, a growing global population, conflicts worldwide, and recent pandemics. Additionally, concerns about food quality, its source, cost, expiry dates, and carbon emissions add further complexity to global food supply chains. In this landscape, blockchain technology emerges as a potential solution, offering the promise of enhancing food authenticity, transparency, and security [55]–[57]. To truly grasp the significance of blockchain technology, let's take a closer look at the entire food supply chain, as depicted in Figure 5 [58]. Here, blockchain becomes a game-changer. Notably, there's a unique solution called the blockchain-enabled credit evaluation mechanism [59]. It uses smart contracts to gather trader data and employs Long Short-Term Memory (LSTM) techniques for data analysis, empowering regulators to make informed decisions that safeguard food supply management [57].

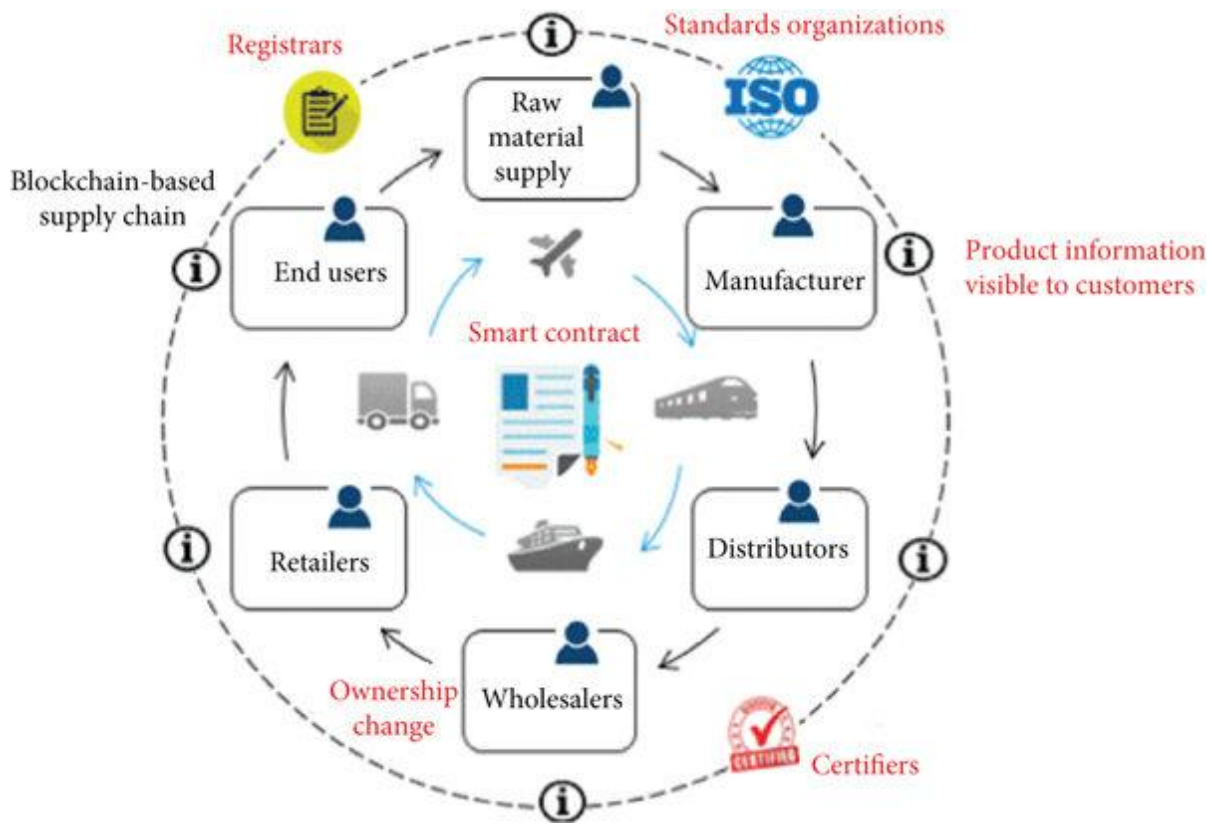

**Figure 5.** Blockchain in the food supply chain

Revolutionizing the traceability of e-commerce products, a groundbreaking blockchain-based approach has been introduced. This pioneering framework [59] hinges on a unique rating-based consensus mechanism known as Proof of Accomplishment. Its primary objective is to bolster traceability and ensure the sustainability of complex supply chains. Exhaustive testing on the Ethereum network has unequivocally affirmed its reliability, delivering both speed and efficiency [60].

Addressing critical issues related to expired food products and traceability, another innovative solution combines IoT and blockchain technologies [61]. IoT devices vigilantly monitor the condition and location of food products, while blockchain technology provides a secure repository for this data. This symbiotic relationship offers an effective solution for identifying unsafe consumables and meticulous control over their journey through the supply chain. Rigorous testing has substantiated the efficiency and resilience of these networks across various configurations and scenarios. In the pursuit of enhancing transparency, security, and trust within the Food Supply Chain (FSC), a comprehensive end-to-end mechanism has been established [62]. Blockchain technology plays a central role, in streamlining data access, mitigating risks associated with single-point vulnerabilities, and introducing embedded TinyML devices to ensure the safety of food monitoring systems [63].

Supply chain management receives a transformative boost as RFID technology is combined with a decentralized blockchain framework [64]. RFID captures data at every stage, from product inception through warehousing to final distribution. This rich tapestry of data seamlessly integrates with the decentralized blockchain. Users gain the confidence to verify connected data blocks during RFID tag scans, instilling data integrity and enabling write transactions on the blockchain. Further exploring the potential of blockchain, a deep dive into the rice supply chain reveals a system fortified with encrypted algorithms, a Byzantine fault-tolerance consensus protocol, and the robust Hyperledger Fabric platform [64]. This innovation leads to a secure and efficient system that rigorously oversees each link within the supply chain.

A comprehensive study addressing a spectrum of issues in the food industry, including safety, waste reduction, loss prevention, and food fraud, has been conducted [38]. Blockchain technology takes center stage in this discourse, offering potential applications. The implications, limitations, and challenges it presents are meticulously dissected, ultimately leading to a range of adaptable solutions for the multifaceted challenges within the food industry. Innovative solutions continue to revolutionize the landscape of supply chains across various domains. This intricate model places a strong emphasis on elevating food quality management across diverse operational layers, spanning from business processes to practical application implementations [16]. In the realm of sustainable tea supply chains, this approach significantly enhances transparency in transactions, fosters price stability, refines the precision of supply and demand forecasts and elevates product safety [65].

Shifting the focus to oil and grain supply networks, this holistic architecture effectively eradicates data uncertainties within supply chains. It has been rigorously validated on the open-source Hyperledger Fabric platform, promising a sustainable and secure future for oil and grain supply networks [66]. In this specific case study involving a retailer and two manufacturers, their research underscores the substantial performance improvements, particularly for products requiring a high level of transparency. In a parallel endeavor, their exploration places a strong emphasis on real-time data collection and transparent accessibility for stakeholders, culminating in heightened traceability, more precise pricing predictions, and an enhanced level of product safety [67]. A different facet of supply chain innovation is presented by this comprehensive model offering a deep dive into the multifaceted factors that shape the tea supply network, providing transparency that empowers consumers to easily verify tea traceability [68]. The authors in this ingenious approach significantly enhance transaction processing capacity while assuring the secure storage of public data through the robust SHA256 [61] algorithm. It also empowers consumers with convenient verification capabilities through QR code scans [69].

Exploring the synergy between IoT and Blockchain within grape wine supply chains, a recent study [70] devised an architecture aimed at streamlining product traceability, enhancing supply and demand management, and eliminating intermediaries and price manipulation. This innovative approach instills greater confidence in the industry and bolsters the integrity of the supply chain. Stepping into the agri-food industry in Iran, a pioneering maturity assessment model tailored specifically for evaluating the readiness of partners to embrace blockchain technology is introduced in [71]. A groundbreaking framework built on consortium blockchain and smart contracts emerges in [46], with a strong focus on enhancing agri-food traceability and data sharing. This forward-thinking approach reduces reliance on intermediaries, optimizes supply timing, and offers easier access to information about product origin and traceability. Data security is fortified through the use of the InterPlanetary File System (IPFS) [72], while a QR code scanner simplifies data access for agricultural products.

A unique approach that integrates reinforcement learning with heuristic search techniques, custom-crafted for the adoption of blockchain in managing autonomous vehicle supply chains is introduced in [73]. This forward-looking approach intelligently navigates the complexities of data traffic, outperforming traditional heuristic methods. In another study, their model's efficacy is rigorously validated using real-world data sourced from the Nosh app [74]. Leveraging the power of Deep Reinforcement Learning, their protocol excels in decision-making for product storage and profit optimization, surpassing Q-learning and heuristic techniques in simulation scenarios [75]. Lastly, this framework, with a strong focus on scalability and efficiency, has undergone rigorous testing, confirming its sustainability, effectiveness in risk management, and viability, particularly in the context of maritime cargo management.

### 4.2 TOURISM INDUSTRY

Blockchain technology has emerged as a game-changer in the realm of smart tourism, offering the promise of eliminating intermediaries and bolstering trust and transparency. Figure 6 vividly illustrates the blockchain can seamlessly integrate with the hospitality industry, benefitting suppliers, business owners, consumers, and policymakers [76]. By eradicating intermediaries, it streamlines interactions among these stakeholders and enforces clear regulations within the tourism sector, ensuring fair billing practices [65].

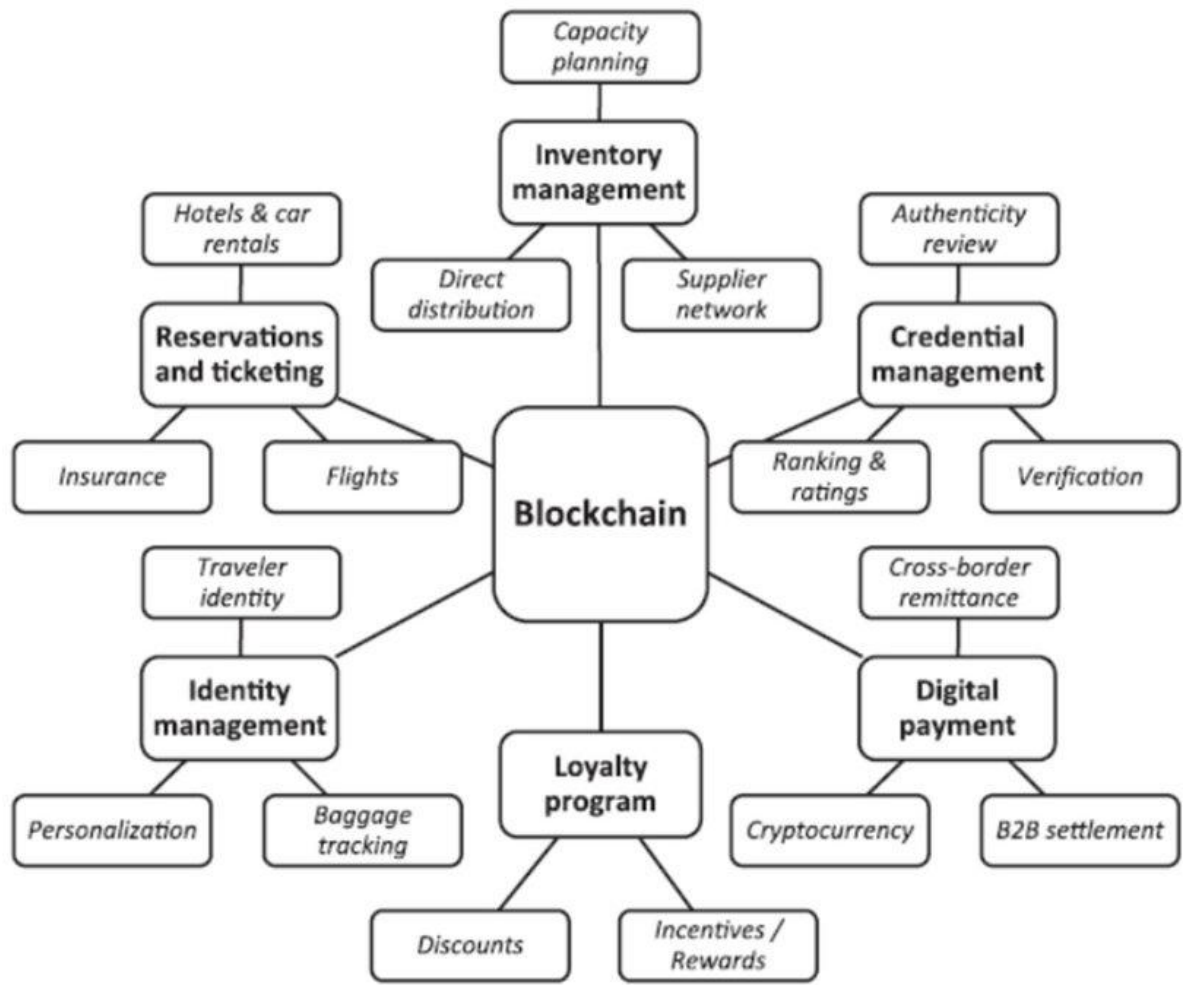

**Figure 6.** Overview of key blockchain applications to enhance tourism

Blockchain has a huge potential in the tourism industry [60]. However, despite its potential, blockchain faces certain challenges when integrating with the sustainable tourism industry. Notably, the 'lack of technical maturity' stands as a prominent obstacle [62]. Efforts such as the BloHosT platform [77] have pioneered integrated services within tourist destinations. This innovative platform employs blockchain for payment purposes, utilizing DL techniques to rate users' experiences. An example of such work in the tourism domain is [78], which presents an innovative blockchain-assisted initiative designed to foster trust and efficiency among various stakeholders and clients. Its explanations employ self-explanatory models and algorithms for clarity.

A recent study [66] addressed data uncertainty in oil and grain supply networks by developing a framework that combines traceability data, a master–slave multi-chain storage system, and link identification models. This innovative architecture has been successfully validated using the open-source Hyperledger Fabric platform [79], promising a sustainable and secure oil and grain supply network [66]. Another study [80], in the pursuit of smart tourism, applied a theme analysis approach to process data and provide a basis for developing a model. The findings advocate for the potential benefits of integrating blockchain and the opportunities and trust among stakeholders [80].

An innovative blockchain-assisted platform [81] offers tourism-related services and professional certification. This architecture, slated for integration into the metaverse platform, not only enhances online travel experiences but also provides users with the ability to earn Non-Fungible Tokens (NFTs) as a form of recognition or reward for their engagement with the platform. These NFTs may represent various achievements, certifications, or virtual assets tied to their tourism-related activities, consequently boosting the platform's native token value [81]. They are unique digital assets stored on a blockchain to represent proof of authenticity or ownership for specific items, often used for digital art, collectibles, and virtual goods. The simulations suggest that blockchain-based platforms are poised to outperform Online Travel Agencies (OTAs) by capitalizing on several key advantages, including the growing familiarity with decentralized technologies among users, the ability to offer more competitive discounts through streamlined processes, the efficiency gained from disintermediation, and the potential emergence of strong industry brands associated with reliability and innovation in the travel sector [82].

A recent study [83] underscored the potential of ICT solutions including Blockchain in tourism and hospitality. Despite the challenges in blockchain adaptation, these technologies hold untapped potential that deserves exploration. Another study [84] explored how blockchain and big data contribute to the cultural and tourism sectors in China. Recent years have seen substantial growth in these sectors by eliminating third-party intermediaries and enhancing sustainability in the tourism and hospitality industry [85]. This architecture promises transparency, data security, and adaptability within the realm of smart tourism. Surveying the implementation of blockchain in Spain's hospitality industry, they uncover a notable gap in its adoption. Despite this, industry professionals recognize the potential contributions of blockchain technology to the sector [86]. Their framework and qualitative techniques advocate for integrating blockchain into the hospitality and tourism economy. This integration is envisioned as a means to empower tourists by optimizing costs, travel times, and check-in processes at accommodations [87]. They offer an architecture that aids travelers in managing their data and facilitates, optimized air travel tickets, tourist destinations, access to accommodation services, and route recommendations [88]. This multifaceted architecture encompasses smart contracts, consensus mechanisms, tokens, and a governance model [89].

A six-parameter-based platform assesses the success of various tourist destinations and stores the data securely on the blockchain, accessible to tourists via their smartphones [90]. By adopting blockchain technology, the tourism industry is poised for transformation, promising greater trust, transparency, and efficiency across the board. These academic endeavors exemplify the industry's commitment to harnessing blockchain's potential for a brighter and smarter future in tourism.

### 4.3 DRUGS MANAGEMENT AND THE HEALTHCARE INDUSTRY

The healthcare and pharmaceutical sectors grapple with numerous challenges, and the dynamic duo of AI and blockchain technology has emerged as a potent solution [84][85]. A prime concern is transparency throughout the drug supply chain, where the infiltration of counterfeit goods poses a grave threat. This nefarious issue has been a focal point for [93], who champion the merits of four-layered blockchain models in clinical practices and precision medicine. Their innovation extends to the creation of the HDG mobile application, a sophisticated tool that likely leverages blockchain for secure and transparent medical record storage. This application not only streamlines healthcare processes but also ensures the confidentiality and integrity of sensitive patient information, marking a significant advancement in the realm of digital healthcare solutions. Through the integration of blockchain technology, the HDG mobile app addresses the pressing need for privacy in medical records while providing a seamless and efficient platform for healthcare management [94].

Innovative methodologies have surfaced in recent research, offering promising applications of blockchain in the healthcare and pharmaceutical landscape. An Ethereum blockchain-based approach [95] leverages smart contracts and decentralized off-chain storage to efficiently trace products in the healthcare supply chain. The smart contract, a linchpin in this framework, upholds data provenance and reduces reliance on intermediaries. This blueprint bestows each involved entity with a secure and immutable transaction history.

A decentralized and multicenter trial, evaluating the real-world feasibility of a blockchain-based dynamic consent framework, is embarked upon in [45]. This ambitious evaluation entails three visits and two follow-up visits for 60 subjects. Subjects are tasked with reporting self-measured body temperatures and virtually administering investigational medicine using unique drug codes within an application. Across the globe, a 3-month qualitative study conducted in Nigeria explores blockchain's potential within the Nigerian pharmaceutical supply chain, aiming to curtail the influx of counterfeit drugs [96].

Blockchain and the IoT to enhance pharmaceutical supply chain traceability are combined in [97]. This distributed ledger maintains an immutable record of all transaction data, ensuring transparency and eliminating the possibility of falsification. Similarly, to harness blockchain to ensure patients receive the correct medications, empowering all stakeholders to verify drug authenticity [98]. Employing the Hyperledger Fabric platform [58], they facilitate peer-to-peer distributed applications for the drug supply chain. To tackle the same challenge, utilizing blockchain-based methods for counterfeit drug traceability within the supply chain [99].

The Medledger system, based on blockchain and chain codes leveraging the Hyperledger Fabric platform, is introduced [100]. This innovative system ensures secure and efficient pharmaceutical transactions among diverse stakeholders. It offers a supply chain management system employing a developed blockchain network, bolstered by role-based authorization to safeguard data integrity [100]. Furthermore, a recent work [101] discussed sub-factors, including trust, government support, international regulation, and industry standards that influence the adoption of blockchain in the pharmaceutical supply chain to combat counterfeiting and investigate the demand for digital drug traceability.

A transparent supply chain underpinned by distributed ledger technology and IoT [102] meticulously assesses scalability and efficiency. Similarly, a hypothetical yet promising model [103] combines IoT and lightweight blockchain techniques to ensure drug traceability. It also advocates for a secure and scalable blockchain approach to drug traceability. Additionally, it delves into digital supply chain management, integrating AL, ML, and blockchain. Meanwhile, pressing issues within the drug supply chain and outlines of how blockchain technology can offer solutions are expressed in [93].

The battle against drug counterfeiting remains at the forefront, as evidenced by [104]. Their model, rooted in blockchain technology, meticulously records drug distribution criteria from manufacturing to patient consumption, bolstered by secure blockchain architecture. Meanwhile, it champions the adoption of blockchain technology to combat drug counterfeiting, implementing its solution using NodeJS and the Hyperledger Fabric platform [47]. This critical issue also finds its place in a book chapter authored by [104], highlighting blockchain's role in mitigating drug counterfeiting.

A blockchain-based drug tracing model [105] detects falsified medicines. It identifies anomalies and substandard drugs from manufacturer to consumer, aided by QR code scanning available on smartphones. A private Ethereum blockchain [106] manages medication control and enhances data security, provenance, transparency, and accountability via smart contracts. The authors further introduce algorithms to elucidate the intricacies of each phase. The pharmaceutical and healthcare sectors stand on the precipice of transformation, with blockchain technology poised to revolutionize transparency, security, and efficiency. These pioneering studies exemplify the industry's commitment to harnessing blockchain's potential, offering a brighter, safer future for pharmaceuticals and patient care.

## 4.4 TELECOMMUNICATION

The integration of blockchain into IoT and 5G/6G networks is an emerging and transformative trend. It's driven by blockchain's inherent qualities of security and transparency, which are poised to revolutionize various aspects of smart cities [85]. These include applications in smart healthcare, industrial IoT, smart transportation, communication with Unmanned Aerial Vehicles (UAVs), and the management of smart grids. Furthermore, the combined utilization of blockchain and UAVs promises to play a pivotal role in enhancing smart communication and healthcare systems.

An energy-efficient dynamic clustering routing algorithm [107] effectively organizes resources in IoT devices. This pioneering technique leverages a self-organizing neural network to establish clusters within the network. Its real-time event detection and node clustering capabilities are assessed using a test-bed study, employing TinyOS. UAVs, or drones, face a multitude of cybersecurity challenges, ranging from GPS manipulation to communication control and jamming [108], [109]. Blockchain technology is poised to efficiently address these challenges, ensuring the safety and reliability of UAV network communication. It achieves this by guaranteeing data immutability and transparency. Additionally, blockchain holds great potential for enhancing UAVs' autonomous control and management. Through the joint utilization of AI, ML, Deep Reinforcement Learning (DRL), and blockchain, sustainable, reliable, and secure UAVs-autonomous vehicle transport systems can be realized [110]–[112].

Substantial progress has been made in the development of a blockchain-based architecture specifically tailored to address the business challenges associated with unsolicited commercial communication. This innovative platform is geared towards elevating subscriber experiences, cultivating fresh opportunities, and strengthening the telecommunications sector as a whole. Its core objective is to enhance data security, safeguard privacy, and optimize operational efficiency throughout the network [106]. A blockchain-enabled architecture [113] ensures equitable resource allocation and task offloading by employing DRL-based smart contracts to optimize handling the highest number of user requests and improve fog revenue. All operations within this architecture are meticulously recorded in blockchain ledgers, thereby enhancing various Quality of Service (QoS) parameters, such as computational energy and latency.

A blockchain-assisted mechanism for secure sharing and access to user data is proposed in [114]. This architecture incorporates a text encryption mechanism, IPFS, and dynamic access control. The utilization of IPFS likely enhances data accessibility and availability by utilizing a decentralized and distributed file system, complementing the blockchain's security features in ensuring a robust and resilient platform for sharing and accessing user data. By storing encrypted data securely over IPFS, access is strictly controlled, ensuring data privacy and security. An architecture that gathers customer data and utilizes ICT to predict proactive re-engagement with clients has been developed [115]. The predictive model employs Support Vector Machines (SVM) and Recurrent Neural Networks (RNN) to identify clients at risk of churning. This groundbreaking framework's efficiency is meticulously evaluated using various metrics, including churn prediction and confusion matrices.

## 4.5 ENERGY MANAGEMENT

Navigating the intricate landscape of renewable energy sources spread across decentralized systems is a substantial challenge within the conventional confines of centralized IT grid frameworks [48]. However, there's a ray of hope in the form of blockchain technology, which has emerged as a robust solution to tackle these complex energy management hurdles. In the realm of renewable energy, blockchain is gaining momentum swiftly, with numerous startups in the global energy industry actively exploring its transformative potential. Blockchain has impressively demonstrated its ability to optimize thermal energy systems, covering aspects such as heating, cooling, and advanced optimization techniques, ultimately resulting in a significant enhancement in overall system performance [116]. Further research underscores blockchain's effectiveness in addressing critical challenges, including digitization, decarbonization, and decentralization in the energy sector [61].

Maintaining equilibrium in modern power systems is of paramount importance. A recent work [117] adopted secure energy management for the operation and optimal scheduling in smart cities. It addresses operational uncertainties by leveraging an unscented transformation-based stochastic architecture. It employs blockchain to ensure the secure transfer of data within the smart city ecosystem.

A secured blockchain-based infrastructure is presented in [118] that manages and analyzes renewable hybrid microgrids in distributed multi-agent environments. It segments the hybrid microgrid into multiple agents and subsequently applies a distributed formula to address optimal planning challenges. The approach has been validated for its feasibility and efficiency [51].

A vehicle-to-everything energy management and power trading system [52] uses IoT, blockchain, and AI technologies. It uses smart contracts to facilitate power trading, matching, and bidding. It seamlessly integrates both local and cloud-distributed ledger nodes to record real-time transaction details. The amalgamation of AI and IoT in communication protocols holds the promise of enhancing green power utilization efficiency, reducing microgrid operating costs, and minimizing power losses.

An energy scheduling framework that operates seamlessly within individual microgrids and extends its reach across multiple microgrids is introduced in [43]. This framework incorporates electric vehicle energy scheduling into microgrid operations, significantly enhancing reliability and security through the adept use of blockchain technology. Validation experiments conducted on an IEEE 118 bus feeder, which encompasses five microgrids, further underscore the remarkable effectiveness of this approach.

The integration of blockchain technology to ensure optimal energy management within transportation automation is explored in [43]. They harness the capabilities of electric vehicles in conjunction with IoT sensors to gather crucial information such as location, distance, and charging levels. This data is meticulously processed through an information hub and subsequently employed in the power scheduling algorithm to determine optimal charging locations and time slots for electric vehicles.

The rise of peer-to-peer energy management, as proposed by [66], offers an empowering solution within renewable energy microgrids. This mode of energy management leverages a blockchain that is custom-tailored specifically for renewable energy microgrids. The framework employs entity mapping to assign unique identities to devices, natural persons, or enterprises, thereby ensuring that only eligible participants gain access to the microgrid. This system adeptly addresses communication delays and fervently promotes a plug-and-play approach. These applications underscore the immense versatility and promise of blockchain technology in revolutionizing the renewable energy sector and energy management systems at large, with a primary focus on enhancing efficiency, security, and sustainability.

## 4.6 SMART TRANSPORTATION

Smart transportation has numerous applications, including automated road speed enforcement, real-time parking management, collision avoidance alert systems, electronic toll collection, and planning and managing traffic [54]. However, it suffers from security and privacy challenges stemming from insecure communication among entities over public channels. Such issues can be addressed by employing blockchain [74]. There is a compelling need for an efficient and lightweight security framework to safeguard the data [16].

A recent study [39] uses blockchain for secure communication in smart transportation. The proposed framework ensures robust access control and key management among various entities, spanning from vehicle-to-roadside unit, vehicle-to-vehicle, and roadside unit-to-cloud server. The authors rigorously evaluate the security of their system, showcasing its resilience against a spectrum of potential attacks. Another study [119] used blockchain for smart transportation to facilitate information sharing without reliance on a 'trusted' data silo. Their strategy, rooted in decentralized systems, addresses four critical requirements: access control, persistence, data integrity, and confidentiality. Smart transportation systems often involve communication among diverse entities, such as pedestrians, vehicles, fleet management systems, and roadside infrastructure over open channels. This scenario exposes vulnerabilities to active or passive interception, deletion, or modification of information during transmission. These security issues and reported solutions are discussed in [120]. A responsive and lightweight methodology for smart transportation systems, leveraging blockchain for authentication and benefiting from the improved capabilities of fog computing over cloud computing to ensure a secure transportation ecosystem is presented in [121].

Securing communication among internet-of-vehicle nodes within smart transportation is focused on [122]. They model their architecture by combining digital twins with big data [123] and employing immutable and traceable blockchain data. Additionally, the authors introduce a privacy-preserving group authentication technique to manage the growing number of users, potentially enhancing access control accuracy and response efficiency. Securing smart transportation systems against automobile hacking is another critical concern, extensively discussed in [124]. Their proposed methodology is grounded in an enhanced security system tailored for self-driving cars, considering aspects like imperial password use, man-in-the-middle attacks, and availability attacks.

Efficient shipment or cargo management plays a pivotal role in smart transportation systems, as exemplified by [125]. They integrated IoT sensors, blockchain-based smart contracts, and ultra-high radio frequency to produce an automatic payment and approval framework for ensuring secure transactions between parties [126]. A model, named SmartCoins, employs consortium blockchain to exploit message validity for enabling vehicles to rate the message source vehicle [127]. SmartCoins are then credited to the vehicle's account on the blockchain, and redeemable at service stations, electric charge stations, and fuel stations. Employing robust cryptographic algorithms such as ECDSA-192 bit [72] and SHA256 [128] ensures message integrity, vehicle authentication, and secure communication. The framework strives to create a transportation system free of fraudulent messages, reduce road accidents, alleviate traffic congestion, and enhance social welfare.

Efficient and secure data sharing necessitates a decentralized network for internet-of-vehicles, prompting [129] to define a fast, decentralized, robust, and scalable fog computing framework. This architecture optimizes vehicular communication and positioning by processing data from the cloud to fog nodes, ultimately enhancing service quality while minimizing network traffic to the cloud server. A similar approach is adopted by [130], utilizing blockchain for authentication within a responsive architecture tailored for smart transportation. The study by [131] introduced an intrusion detection system that detects attacks by employing federated deep learning. Specifically, it employs a context-aware transformer mechanism to learn spatial-temporal representations of potential flows in vehicular traffic, aiding in the classification of different attacks. Blockchain is integral to their framework, facilitating reliable, distributed, and secure training across multiple edge nodes.

Block-CLAP, a certificates key [132] agreement framework employing blockchain for the Internet of Vehicles (IoV) is introduced in [133]. Traffic-centric data within the network is securely transferred to a cluster head and neighboring roadside units using established secret keys. The cloud server collects this information, generates transactions, and forms blocks, with blockchain verification and inclusion via a voting-based consensus method. Blockchain converges with Digital Twins and predictive analysis tools [123] to address the collaboration requirements of digital twins. This architecture incorporates real-time operational data analytics and distributed consensus. The proposed model is tested across various use cases, including smart logistics, smart transportation, and railway predictive maintenance, demonstrating its feasibility.

Data vulnerability within a sustainable smart city is a significant concern, tackled by [134] through an IoT and blockchain-based decentralized data management network. The authors employ prior knowledge to establish a hyperledger fabric-based data system, fostering a smart, secure, and trusted transportation network. The use of IoT with blockchain led to an automated rescue and service provider model for highways [135]. This system includes an intelligent vehicle deviation component, aimed at reducing traffic delays, saving time, and minimizing fuel consumption [126].

A recent work [136] integrated ML, AI, and blockchain to produce a decentralized network for mitigating security issues in smart transportation. It proposed A blockchain-based quantum approximate optimization method to enhance scalability and reduce costs in smart logistics, ensuring smart logistic network security and a secure distributed ledger [136].

A shared mobility platform [21] employed blockchain with ERC-721 tokens [137], hardware security blocks, and smart contracts to safeguard confidential key material and thus foster secure mobility services. Blockchain is positioned as a trust infrastructure for smart city systems [138]. The authors categorized the database as authorized access, shared, or private and designed blockchain architectures as private, public, or consortium. They also established selection criteria to authenticate blockchain nodes and mechanisms for dynamic smart blockchains, promoting interoperability. A cost-effective privacy-preserving mechanism, centered around pseudonym management, and based on blockchain is proposed in [23] for smart transportation. This approach particularly focuses on protecting the position and identity of vehicles to prevent tracking by malicious entities.

## 5 FEDERATED LEARNING AND APPLICATION IN SUSTAINABLE SMART CITIES

Federated Learning (FL) is an innovative machine learning approach where the global algorithm is executed on individual devices, avoiding the need to transfer raw data to a central model [139]. Parameters from local devices are sent to the central model for training and predictions [140]. This decentralized method addresses challenges in smart cities, offering solutions for privacy preservation and efficient handling of big data, enabling real-time decision-making [141]. FL facilitates collaborative model training without sharing sensitive data, enhancing privacy in machine learning applications. Figure 7 summarizes graphically the applications of federated learning in developing and maintaining sustainable smart cities [142].

### 5.1 ENHANCING SMART TRANSPORTATION SYSTEMS

Smart transportation is all about seamlessly integrating cutting-edge technologies to ensure cost-effective, efficient, and secure traffic management. It leverages advanced systems to monitor, analyze, and optimize transportation networks, ultimately boosting their operational efficiency and safety [75], [143]. Some of the key applications are listed below [142]:

- *Traffic Flow Optimization*: By harnessing data from diverse sources, including sensors, traffic cameras, GPS-enabled devices, and sensors, the FL models can predict real-time traffic patterns and congestion points. This invaluable information empowers traffic authorities to make timely adjustments such as signal timing, vehicle rerouting, and congestion mitigation.

- *Predictive Maintenance*: By continuously collecting data on vehicle performance, FL algorithms can forecast maintenance needs, effectively reducing downtime and enhancing overall transportation efficiency.
- *Public Transport Scheduling*: By analyzing passenger demand patterns, the FL models fine-tune schedules to match peak hours, thereby minimizing passenger waiting times and enhancing the overall public transport experience.
- *Parking Management*: When integrated with sensors and data from parking lots, FL algorithms assist drivers in quickly locating available parking spaces, ultimately reducing traffic congestion and lowering carbon emissions.
- *Safety Enhancing:* By analyzing real-time data from vehicles and infrastructure, the FL models can identify potential safety hazards, promptly alerting both drivers and authorities to take preventive actions.
- *Emission Monitoring and Control:* The integration of FL enables real-time monitoring of vehicle emissions, aiding in the assessment of air quality and pollution levels, and enabling policymakers to implement measures for reducing emissions and enhancing urban air quality.
- *Route Optimization:* FL can optimize routes for individual vehicles by exploiting real-time traffic data and historical patterns, resulting in reduced travel time, fuel consumption, and greenhouse gas emissions.
- *Traffic Management:* By processing data from various sources such as sensors and traffic cameras, the FL models enable dynamic management of traffic signals and lane assignments, which minimizes congestion and improves the overall traffic flow.
- *Public Transport Demand Forecasting:* The FL models can assist in efficient resource allocation and optimization of public transport services by predicting public transport demands based on historical usage and external factors like events and weather conditions.
- *Pedestrian and Cyclist Safety:* The FL models can enhance the safety of pedestrians and cyclists by analyzing data from various sources to identify potential collision risks and unsafe areas.

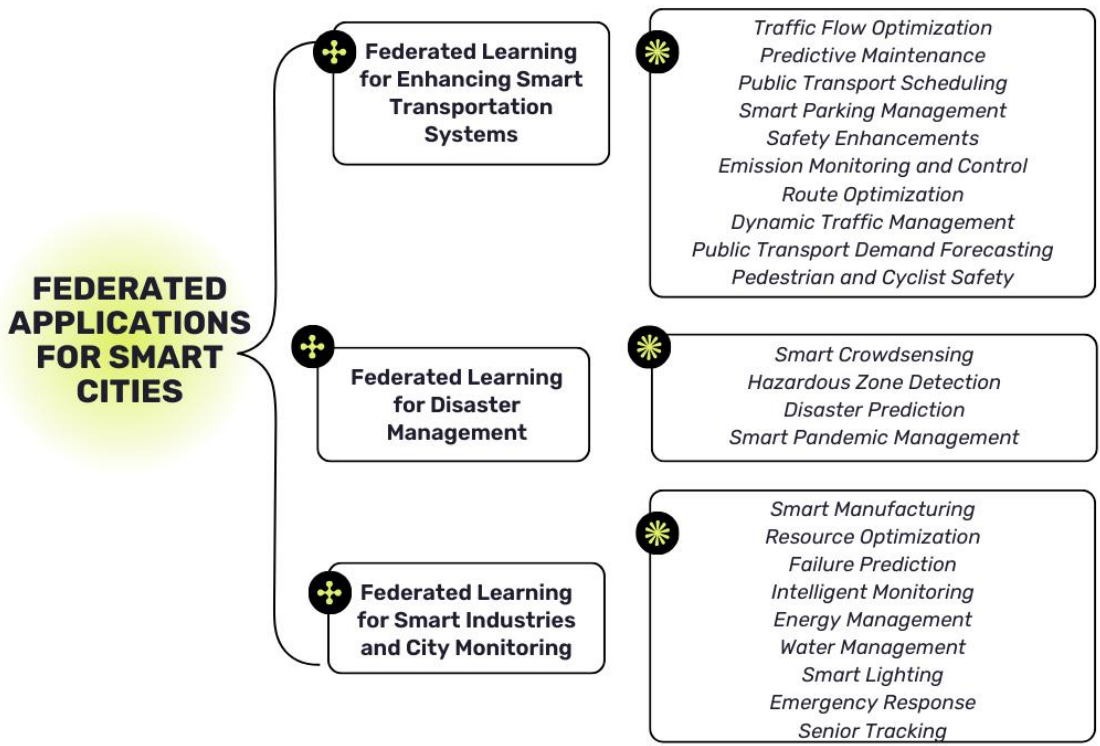

**Figure 7.** Federated Learning Applications in Sustainable Smart Cities

## 5.2 DISASTER MANAGEMENT

Effective disaster management, whether it's responding to earthquakes, floods, cyclones, or pandemics, is a critical aspect of ensuring the safety and well-being of communities. Smart disaster management [144] leverages cutting-edge paradigms and technologies to achieve several key objectives: (a) minimizing response times during disasters; (b) promptly monitoring public health, especially during pandemics; and (c) making informed decisions for efficient disaster management [145]. This is where FL steps in as a valuable ally. How FL can enhance disaster management within the context of smart cities is discussed as follows [146]:

- *Smart Crowdsensing*: Smart crowdsensing harnesses the power of embedded sensing technologies in mobile devices to meet specific Quality of Service (QoS) requirements. Autonomous crowdsensing can be rapidly deployed at reduced costs, significantly enhancing the responsiveness of disaster management efforts. FL ensures that the insights gathered from these devices are collectively utilized without compromising user data privacy.

- *Hazardous Zone Detection*: By applying advanced image vision techniques to aerial images, FL models can accurately identify hazardous zones with minimal human intervention. This allows for swift and precise mapping of danger areas, enabling efficient evacuation and resource allocation.
- *Disaster Prediction*: FL, driven by AI-driven risk analysis, correlates historical disaster events with current conditions. This correlation aids in predicting future disasters, thereby providing critical lead time for preparedness and mitigation efforts. By securely aggregating data from various sources, FL contributes to more accurate risk assessments and early warning systems.
- *Smart Pandemic Management*: During pandemics like COVID-19, FL can be instrumental in monitoring and managing the crisis [147]. FL-powered techniques can predict health risks, and death rates, and assist in efficient resource allocation. These strategies, driven by FL's privacy-preserving capabilities, enhance healthcare planning and ultimately save lives while safeguarding sensitive health data [87], [148]

## 5.3 IMPACT ON SMART INDUSTRIES AND CITY MONITORING

The emergence of smart industries, driven by Industry 4.0 technologies, has ushered in a new era of manufacturing efficiency, resource optimization, and predictive maintenance. Simultaneously, smart buildings harness the power of the IoT and data analytics to revolutionize energy management, security, and communication within cities. Federated learning seamlessly integrates into these domains, enhancing their capabilities in several ways [149], discussed below:

- *Smart Manufacturing*: Federated learning enhances smart manufacturing by allowing distributed AI models to learn from data generated across various factory locations, facilitating the optimization of industrial processes, improving efficiency, reducing errors, and minimizing the need for extensive human intervention.
- *Resource Optimization*: FL contributes to resource optimization within smart industries by enabling collaborative model training. This reduces communication and computation costs while minimizing the environmental impact of resource-intensive processes.
- *Failure Prediction*: Predictive maintenance systems in smart industries benefit from FL's ability to train models on diverse datasets without centralizing sensitive information. These systems forecast potential glitches in production lines, allowing for proactive maintenance, minimizing costly failures, and improving overall system robustness.

- *Intelligent Monitoring*: In smart buildings, FL offers distributed AI models to analyze CCTV images for private spaces and enhances security by detecting potential threats while reducing energy waste through efficient lighting and climate control.
- *Energy Management*: By utilizing IoT and data analytics, FL models can optimize energy consumption patterns based on real-time data, leading to reduced environmental impact, improved energy efficiency, and cost savings.
- *Water Management*: By estimating water consumption patterns via an IoT-enabled measurement system, FL can address water scarcity issues in urban environments.
- *Smart Lighting*: IoT-based lighting systems are further optimized with FL, which enables collaborative learning to allow systems to adapt to occupant preferences while simultaneously minimizing energy consumption without compromising user privacy.
- *Emergency Response*: Smart alarm systems, powered by IoT technology, benefit from FL's ability to aggregate data from various sources without centralizing it. This allows for quick notifications and responses to unexpected hazardous conditions while preserving the privacy of residents.
- *Senior Tracking*: FL-enhanced IoT-based tracking systems for senior residents promote their health and safety by monitoring health metrics, and emergency calls, and analyzing behavior patterns discreetly without compromising the privacy and security of sensitive health data.

## 6  NLP Applications in Sustainable Smart Cities

Natural language processing (NLP) stands at the intersection of technology and data, representing a remarkable field that harnesses the power of ML, probabilistic techniques, and statistical methods. Its journey has been nothing short of remarkable, with rapid strides in Computer Science, elevating its stature in both domestic and industrial realms. The past decade has witnessed a dramatic transformation, thanks to the emergence of more efficient neural and DL models. These advancements have not only enhanced precision but have also endowed NLP with newfound versatility in terms of numerous applications, including but not limited to automated machine translation, clinical NLP for improved doctor-patient interactions, voice-based applications (e.g., voice-controlled devices, educational platforms, customer service interactions, voice and multimedia analysis, and the ever-helpful virtual assistants [150]), predicting disease associations [151], sentiment analysis, and controlling, monitoring, and supporting IoT-based applications [81]. Figure 8 offers an illustrative representation of NLP's presence across various aspects of life, underscoring its pivotal role in shaping intelligent systems. The major application of NLP and its role in sustainable smart cities are discussed in the following sections.

## 6.1 SMART HEALTHCARE

Within the healthcare sector, clinical NLP, or C-NLP, takes center stage. This domain is particularly significant due to the abundance of unstructured data, encompassing medical reports, electronic health records (EHRs), tests, imaging data, patient records, and feedback reports. The potential insights hidden within this trove of data, unlocked by the application of AI and ML techniques, are immense. Here, NLP plays a pivotal role in diagnostic research, document management, methodology comparisons, prognostic analysis, and various other medical operations [83], [148]. C-NLP applications span four primary categories:

- *Non-medical Procedures*: Streamlining processes like patient admissions, discharges, referrals, and priority categorization for enhanced efficiency and accuracy.
- *Cancer Studies*: Various aspects of cancer research and treatment, from diagnosing cancer types and stages to predicting reoccurrence and assessing toxicity symptoms [44].
- *Mental Health Allied Research*: Detecting early symptoms of mental health disorders and offering support for patients, with applications in analyzing psychotherapist notes for diagnoses.
- *Miscellaneous Practices*: NLP finds applications in diverse medical domains, including joint infection detection, fracture risk prediction, HIV care retention studies, and addressing emerging diseases like COVID-19.

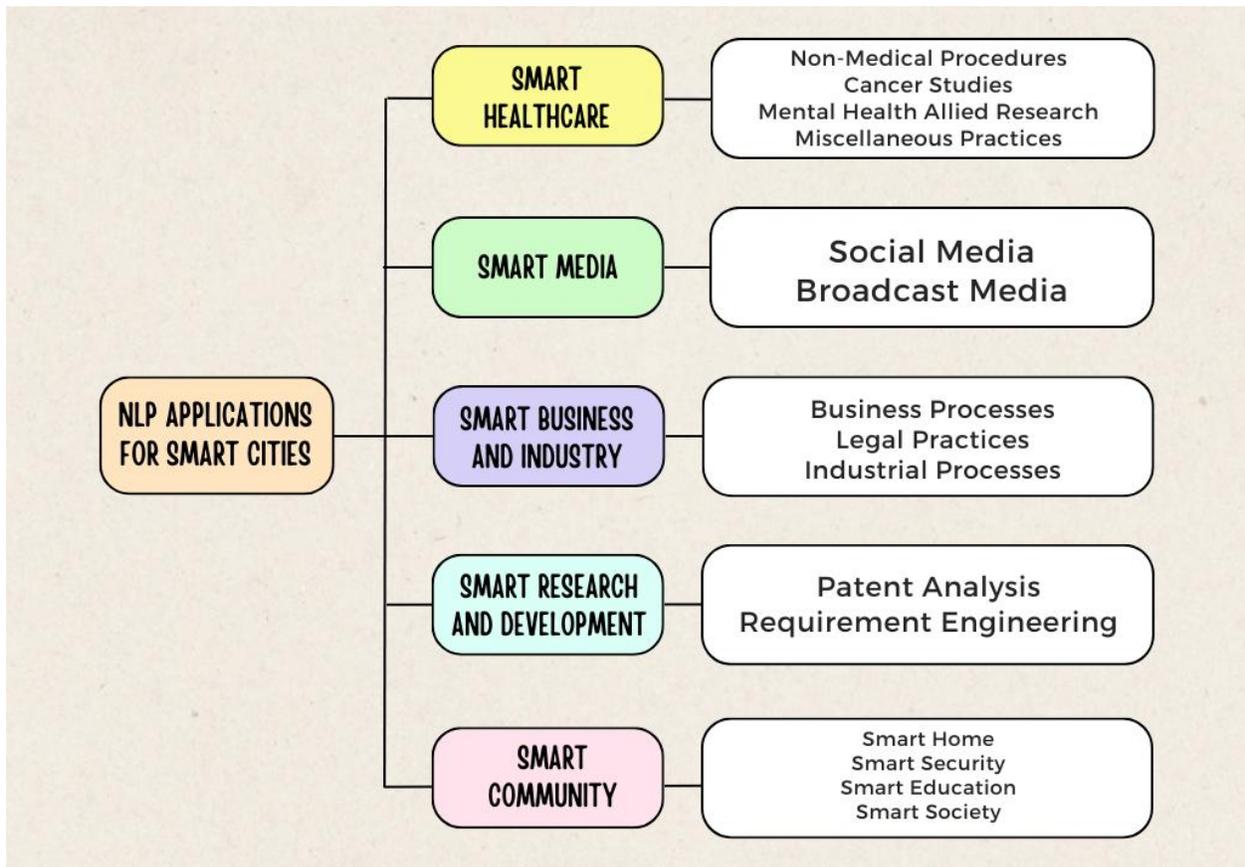

**Figure 8.** NLP Applications in Sustainable Smart Cities

**6.2 SMART MEDIA**

In today's digitalized media landscape, we are inundated with an immense volume of unstructured data, presenting a realm of opportunities for intelligent applications. These applications encompass text summarization, data analysis, rumor detection, caption and headline generation, etc. Within the domain of smart media, the diverse applications of NLP can be elegantly categorized into two distinct realms: those centered around social media and those focused on broadcast media [152].

**6.2.1 Social Media**

The explosive growth of social media platforms has ushered in an era of prolific user-generated content. This necessitates the development of efficient methods for content curation and analysis, where NLP takes center stage, addressing various facets of social media [65]:

- *Content Filtering and Censorship*: NLP tools assume the role of gatekeepers, filtering and censoring content, thereby extracting the essence of meaningful information while suppressing harmful or inappropriate data.
- *Opinion Mining and Sentiment Analysis*: NLP techniques prove instrumental in sifting through the vast sea of textual content to extract sentiments and opinions. This, in turn, enables the analysis of user sentiments towards a myriad of topics and issues.
- *Rumer Detection*: In the age of rampant misinformation and rapidly spreading rumors, NLP emerges as a formidable shield. It detects and counters the unfettered propagation of rumors and falsehoods across online platforms by meticulously analyzing patterns and inconsistencies in textual content.

### 6.2.2 Broadcast Media

This category encompasses traditional media outlets such as newspapers, television news, and public websites, where NLP-driven solutions find their niche in content generation and summarization [22]:

- *Text Summarization*: NLP techniques are harnessed to automatically distill lengthy articles into concise and informative summaries. This empowers users with key insights without the need to delve into the entire expanse of content.
- *Caption and Headline Generation*: NLP's creative prowess comes into play when it comes to crafting attention-grabbing headlines and captions for media content. These captivating snippets enhance user engagement and facilitate the dissemination of crucial information.

## 6.3 Smart Business and Industry

In the vast landscape of corporate giants, small enterprises, stock exchanges, legal firms, industrial magnates, and financial powerhouses, a treasure trove of structured and unstructured data resides. This reservoir of data beckons with opportunities for forecasting, evaluation, analysis, and fortification of decision support systems. At the very heart of this technological revolution stands AI, with its most adept tool – NLP. A lion's share of this data is cloaked in the garb of natural language text, affording NLP as a versatile arsenal for data processing, ultimately culminating in invaluable insights. In this segment, we embark on an exploration of the multifaceted applications of NLP spanning the spectrum of business and industry.

### 6.3.1 Business Processes

In the relentless face of rapid globalization, urbanization, and cutthroat market competition, businesses are called upon to swiftly adapt to technological shifts and evolving landscapes. Organizations are increasingly turning to AI solutions, with NLP leading the charge, to streamline business processes and facilitate data assessment [88].

- *Extraction of Business Terminology and Rules*: Extracting precise business terminology and rules is among the initial and resource-intensive tasks in establishing any business. Danenas et al. [169] proposed an NLP-based tool for extracting the semantics of business vocabulary and rules to expedite the process by utilizing UML diagrams from relevant projects.

- *Predictive Business Process Management*: The ability to anticipate the trajectory of business processes is pivotal for averting failures and unwarranted fluctuations during execution. The POP-ON predictive model [85] uses NLP in business process management for process mining by exploiting log data from various business operations.

- *Financial Insights from Textual Data*: NLP assumes a pivotal role in extracting financial insights from the vast landscape of business news, facilitating stock market and earnings forecasts. An extensive study on NLP techniques [153] emphasizes the efficacy of contextual embeddings and transformers in this context.

- *Legal Document Comprehension*: NLP offers invaluable support for the classification, analysis, and summarization of legal documents.

### 6.3.2 Legal Practices

Legal documents, by their very nature, often comprise intricate textual content, demanding a high degree of comprehension and interpretation. NLP steps in to offer solutions for classifying, analyzing, and summarizing these dense texts, thereby enhancing comprehension and decision-making. Task-specific models, meticulously trained on legal data, serve as the linchpin for these applications.

- *Statutory Reasoning Implementation*: Several BERT-based models have been developed for statutory reasoning. Examples include Legal-BERT trained on the SARA (Statutory Reasoning Assessment) dataset [88] is capable of clearly defining legislative rules. Legal-BERT has been meticulously optimized for legal-NLP tasks [89].

- *Automated Legal Contract Review*: Several works have been done on automatically reviewing legal contracts. An example is the use of transformer-based models trained on CUAD (Contract Understanding Atticus Dataset) have shown impressive efficiency [89].
- *Predicting Legal Rulings*: NLP has also been used in predicting legal rulings. An example is the use of ML classifiers like RF, SVM, and Decision Trees with DL methods such as LSTM and Bi-LSTM to tackle this complex task.

### 6.3.3 Industrial Processes

Within the industrial domains, applications of NLP manifest across a spectrum of functions, including monitoring systems, decision support systems, risk analysis, process prediction, real-time data analysis, resource allocation, command-based automated manufacturing, and data processing [90].

- *Forecasting Virtual Machine Loads*: An adaptive network fuzzy inference system was developed by [91] to predict the virtual machine loads for predicting resource utilization and optimizing resource allocation. It used NLP to extract instructions from command-line applications, which were analyzed using grey relational analysis [91] before inputting the proposed system.
- *Predicting Accidents in the Construction Sector*: In the construction sector, analyzing past hazardous accidents is paramount for preventing future incidents. An example is the use of text mining using NLP and ML algorithms [92] to analyze reports of prior accidents to predict the root causes of these incidents. A similar study employed NLP algorithms to gain insights regarding incidents in the pipeline industry [69]. Another study integrated NLP and random forest algorithms to analyze mine safety and health administration data [95]. The aim was to replace the manual categorization of incidents from these huge dumps of data, which is a labor-intensive undertaking [45].

### 6.4 Smart Research and Development

In this research and development (R&D) landscape, NLP has numerous applications such as patent document analysis and requirement engineering, significantly impacting the trajectory of technological advancement.

### 6.4.1 Patent Analysis

In the realm of intellectual property rights, patents emerge as the custodians of novel technologies, offering protection and strategic advantages to their holders. Effective patent analysis stands as a cornerstone for companies, providing critical insights into market competition and innovation strategies.

- *Semantic Patent Recommendations*: NLP and neural networks are among the cutting-edge technologies to identify and recommend semantically similar patents [96], thereby facilitating trend analysis and technology mining.
- *Simplifying Complex Patent Language*: Patent language, especially within specifications and claims, is often steeped in intricate terminology [97]. The application of NLP to extract core inventive information from patent claims.
- *Automatic Patent Summarization*: The automated summarization of numerous patent documents is of paramount importance. A study [98] introduced a sequence-to-sequence with an attention model to automatically summarize patterns, preserving the essence of this vital knowledge.
- *Identifying Emerging Technologies*: The NLP algorithms can be employed in identifying emerging technologies and themes from patent documents [99].

### 6.4.2 Requirement Engineering

In the dynamic realm of software development, NLP emerges as a powerful ally in the automated analysis of software requirements, especially security-related concerns.

- *Extracting Security Requirements*: The NLP algorithms can assist developers in identifying potential security vulnerabilities. An example is an NLP-based approach [154] that extracts security requirements from software requirement specification documents.
- *Automated Requirements Validation*: NLP and ML algorithms can be employed to automatically validate software requirements by comparing the requirements against industry-specific standards and best practices [95].
- *Efficient Requirements Change Management*: The NLP-based approaches offer valuable support by analyzing the potential impact of proposed requirement changes. A model that leverages NLP to scrutinize the ramifications of requirement changes and recommends appropriate actions to developers has been presented in [100].

- *Automated Test Case Generation*: In agile software development, NLP lends its prowess to the automated generation of test cases from these user stories. An example is the NLP-based algorithm [81] that uses NLP algorithms to seamlessly generate test cases from user stories.

## 6.5 Smart Community

Communities and their residents are essential components of any smart city. Solving challenges at the community level involves catering to diverse demographics and individualized service applications [154]. The synergy of ICT, IoT, and AI has been instrumental in crafting solutions for the development of smart communities. NLP, either independently or in conjunction with IoT, is applied across various domains of smart communities, encompassing education, society, smart homes, and security solutions. This section delves into these domains within the context of smart communities.

### 6.5.1 Smart Home

A smart home represents a living space enriched with intelligent technological aids, augmenting residents' quality of life and streamlining daily tasks [155]. Individuals can personalize their homes with applications that enable control, monitoring, and maintenance of various aspects, promoting independent living. NLP techniques are pivotal in conceptualizing smart homes and developing their diverse features and products [86].

- *Voice-Controlled Home Automation*: A home automation model that seamlessly integrates IoT sensors and NLP-based voice control is introduced in [81]. This system empowers individuals, particularly those with disabilities, to utilize voice commands for controlling door locks, temperature settings, alarm systems, and more. NLP acts as the conduit between users and their smart devices.
- *Voice-Activated Lighting Control*: A system employing Raspberry Pi and NLP-based voice recognition to replace traditional light switches implemented in [100]. Users can remotely manage house lighting using voice commands, benefiting the elderly and differently-abled individuals.
- *IoT-enabled Smart Homes*: A smart home architecture is grounded in IoT, seamlessly integrating Google Assistant for voice control [155]. A Node MCU [135] links IoT devices to Google Assistant, accessible through smartphones, with cloud storage for data processing.

- *Enhancing Home Security*: A smart home system architecture with a focus on improving security and user authentication features presented in [101]. NLP plays a pivotal role in processing natural language commands, fortifying the security of smart homes.

### 6.5.2 Smart Security

Security constitutes a critical facet of smart communities, encompassing both physical and virtual security measures. Physical security entails authentication systems, secure access, and CCTV surveillance, while virtual security centers on safeguarding communication, assets, and online data. NLP technologies contribute significantly to enhancing security measures.

- *Call Pattern Analysis*: An NLP-based clustering method to scrutinize recurrent calls to public safety services is devised in [102]. This approach identifies major call causes and patterns by analyzing unstructured government records, bolstering public safety.
- *Malware Detection*: The menace of malware threats by tracking system calls and detecting malicious applications through NLP is addressed in [156]. They employed innovative techniques such as Bags of System Calls and cosine similarity algorithms [16], [69], [149], [156].
- *Phishing Detection*: A sophisticated model that fuses NLP and graph convolutional networks (GCN) to detect phishing emails has been reported in [104]. While NLP processes the email text, GCN performs text classification to identify phishing content, fortifying virtual security.

### 6.5.3 Smart Education

NLP plays a pivotal role in elevating the teaching-learning process within the ambit of smart education. It assists educators in crafting e-learning platforms, analyzing study materials, generating sample questions, and evaluating teacher-student feedback.

- *Sociopolitical Text Analysis*: Lucy et al. [157] harnessed NLP techniques to analyze the sociopolitical aspects of text within U.S. history books, delving into how specific demographic and cultural groups have experienced marginalization. Their methods included lexicon-based approaches, word embeddings, and topic modeling.

- *Automated Question Generation*: Deena et al. [158] automated the generation of Multiple Choice Questions (MCQ) and subjective questions through NLP and Bloom's taxonomy. This streamlined the creation of question banks for online assessments.
- *Multilingual NLP in Education*: Maxwell-Smith et al. [44] transcribed Indonesian-English classroom speech to create bilingual corpora using automated speech recognition and NLP techniques. This facilitated multilingual NLP processes, enhancing the accessibility of education.
- *Feedback Analysis*: A Systematic Mapping Review (SMR) on NLP systems employed for semantically analyzing student feedback and teacher assessments is conducted in [104]. Following the PRISMA framework, their work provided valuable insights into how NLP enriches the understanding of these forms of feedback.

### 6.5.4 Smart Society

The benefits of NLP techniques extend to broader society, supporting community-level development and addressing challenges arising in smart cities experiencing population growth.

- *Community Profile Information*: Stories2Insights (S2I), an automatic user-perceived value classification model designed to collect community profile information through interviews is introduced in [104]. This invaluable data aids sustainable development efforts.
- *Social Needs Extraction*: NLP to extract social needs information from EHRs is harnessed in [104]. This extraction facilitates the classification of hospital-admitted patients based on their unique social requirements, thus enhancing community-level healthcare delivery.

## 7   LARGE LANGUAGE MODELS

Natural language understanding covers a broad spectrum of tasks aiming to comprehend input sequences better. Initially, the primary objective for developing language models, large language models (LLMs), was to enhance performance in NLP tasks, encompassing both understanding and generation of human language. In recent years, LLMs, including notable examples like ChatGPT, have significantly advanced NLP and related areas. These models have shown emergent behaviors, such as the ability to "reason," particularly when they reach substantial sizes. For instance, by providing these models with a "chain of thoughts" or explicit reasoning exemplars or simply instructing them to "think step by step," they can produce answers to questions with well-defined reasoning steps. This development has piqued considerable interest in the research community, as reasoning abilities are a hallmark of human intelligence that has often been considered a missing element in current AI systems [110], [111], [159].

The Large Language Models have been quite popular in academic and industrial circles due to their exceptional performance across various applications. They are computational models that comprehend and generate human language with the remarkable ability to predict the likelihood of word sequences and generate new text considering the provided input [85],[86]. The most prevalent type of LM, N-gram models [115], estimate word probabilities by considering the context that precedes them.

However, despite their impressive performance in some tasks, there remains uncertainty about the extent to which LLMs are genuinely reasoning and whether they are doing so effectively. To some researchers like [112], these models are still limited in demonstrating acceptable performance regarding planning and reasoning tasks that are much easier for humans. They face various challenges, including handling rare or previously unseen words, addressing overfitting issues, and capturing intricate linguistic phenomena. To overcome these challenges, researchers are continually dedicated to enhancing LM architectures and refining training methods. This section gives a brief overview of the recent advances in LLMs and some of their notable applications.

## 7.1 RECENT ADVANCES IN LLMS

Large language models have been the center of Computer Science and AI, where the research community has been actively involved in their development and has shown tremendous improvements in certain areas. Some of the notable progress are outlined below:

- *Emergent Reasoning Abilities*: Studies by [160] indicate that reasoning abilities appear to emerge in large language models, notably in models like GPT-3 175B. These abilities manifest as significant improvements in performance on reasoning tasks at a particular scale, typically around 100 billion parameters. This suggests that employing large models for general reasoning problems may be more effective than training small models for specific tasks. The underlying reasons for this emergent ability are still not fully understood, and potential explanations are explored in [117].

- *The Influence of Chain of Thought (CoT) Prompts*: The use of CoT prompts, as demonstrated by [117], has been found to enhance the performance of LLMs on various reasoning tasks. It allows LLMs to produce valid individual proof steps, even when dealing with fictional or counterfactual ontologies. However, LLMs may occasionally make incorrect or incomplete proofs when multiple options are available. Moreover, the use of CoT prompts has been shown to improve the out-of-distribution robustness of LLMs, an advantage not typically observed with standard prompting or fully supervised fine-tuning approaches [117].

- *Human Like Content Effects on Reasoning*: LLMs exhibit reasoning patterns similar to those observed in human cognitive processes [161]. The models' predictions are influenced by prior knowledge and abstract reasoning, and their assessments of logical validity are affected by the believability of conclusions. This suggests that although LLMs may only sometimes excel at reasoning tasks, their challenges often parallel those humans find demanding. This implies that language models may "reason" in ways akin to human reasoning.

- *Complex Reasoning*: Despite their impressive reasoning capabilities in specific contexts, LLMs, as indicated by [105], and [112] struggle with more complex reasoning tasks or those involving implicature. These models must be more robust even in simple, commonsense planning domains that humans navigate effortlessly, such as GPT-3 and BLOOM. This suggests that existing benchmarks may need to be more complex to fully assess the true reasoning abilities of LLMs, and more challenging tasks are required for comprehensive evaluations.

## 7.2 LLM APPLICATIONS

This section briefly reviews the application of LLMs in medication and education.

### 7.2.1 Medical Applications

There has been a notable surge in the use of LLMs in the medical field, prompting an in-depth exploration in this section. The efforts to harness LLMs for medical applications are categorized into three aspects, namely medical query, examination, and assistant, as summarized in Table 5.

- *Medical Queries*: LLMs, including ChatGPT, have been successful in answering medical queries spanning genetics, biomedicine, radiation oncology physics, etc. [162]. However, there is still room for improvement, as some evaluations highlight limitations, such as the potential for ChatGPT to generate responses without reliably citing sources and the potential for fabricating information [162].

- *Medical Examinations*: The performance of LLMs has been evaluated in medical examination assessments, including the United States medical licensing examination. Results indicate that ChatGPT achieves varying accuracies across different datasets, with some challenges related to out-of-context information [163]. However, ChatGPT's performance, without tailored training, demonstrates its capability to assist medical practitioners and help in informed clinical decision-making, offering valuable resources and support to medical students and clinicians [164].

- *Medical Assistants*: LLMs show potential in medical assistance, with applications including the diagnosis of dementia, and gastrointestinal diseases, and accelerating the evaluation of COVID-19 literature [44], [165], [166]. These models can provide valuable insights into healthcare. However, they also come with challenges and limitations. These include lack of originality, resource constraints, high input requirements, uncertainty in answers, patient privacy, and the risk of misdiagnosis [43], [87], [148]. Studies show that ChatGPT is performing well and is feasible in clinical and medical education and surgical training. It can enhance various aspects of medical practice. However, further work is needed to address their limitations and fully unlock their capabilities.

### 7.2.2 Education

LLMs show potential in transforming the field of education with applications that can aid students improve writing skills, comprehending complex concepts, expediting information delivery, and boosting their engagement via personalized feedback. The aim is to present an efficient interactive learning environment with lots of educational opportunities for students, where the full potential of LLMs in education demands further research and development. Evaluations of LLMs for educational support seek to examine and assess their potential contributions to the educational field. These evaluations can be approached from various angles. According to [167], ChatGPT can produce detailed, coherent, and fluent feedback on students' assignments, aiding in their skill development by even surpassing human teachers. However, sometimes ChatGPT lacks insightful teaching perspective and novelty. Some studies found that LLMs can pinpoint at least one actual issue in the programming code produced by a student, although misjudgments have also been observed [168]. Therefore, further research is required to achieve proficiency in addressing logic issues, and output formatting, and generate fewer errors similar to those made by students [169].

Researchers have evaluated how LLMs are effective in educational examinations, such as generating questions, scoring answers, and guiding the learning process. In this regard, ChatGPT achieved the average 71.8 correctness score against the average score of all students participating in the contest. Subsequently, evaluation using GPT-4 produced a score of 8.33 [170], showing that bootstrapping, where randomness is combined with the "temperature" parameter led to incorrect answers [171].

## 8 CHALLENGES IN SUSTAINABLE SMART CITIES

Starting up the smart urban areas is a daunting and challenging task, involving several unavoidable factors including cost, proficiency, maintainability, correspondence, well-being, and security. The best cluster in making these huge models is cost, which can be due to planning, activities, configuration, efficiency, manageability, carbon release, city wastage expulsion, effective working expenses, natural disasters, provision of ICT, sensors, and IoT gadgets, force disappointments, data protection, security, and privacy. This section portrays some of the challenges in developing sustainable smart cities. Figure 9 graphically illustrates some of the challenges faced by smart cities before and after their creation. Figure 10 presents the general challenges faced by smart cities.

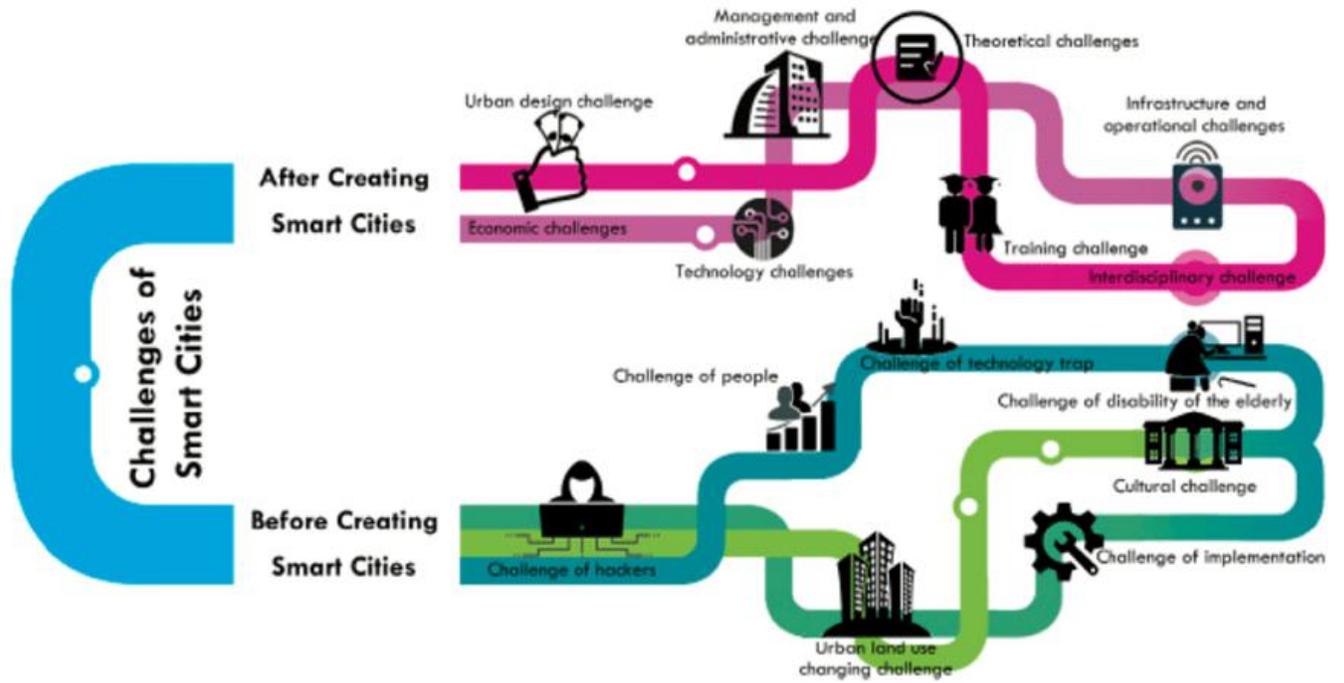

**Figure 9.** Challenges of Smart Cities before and after their creation [172]

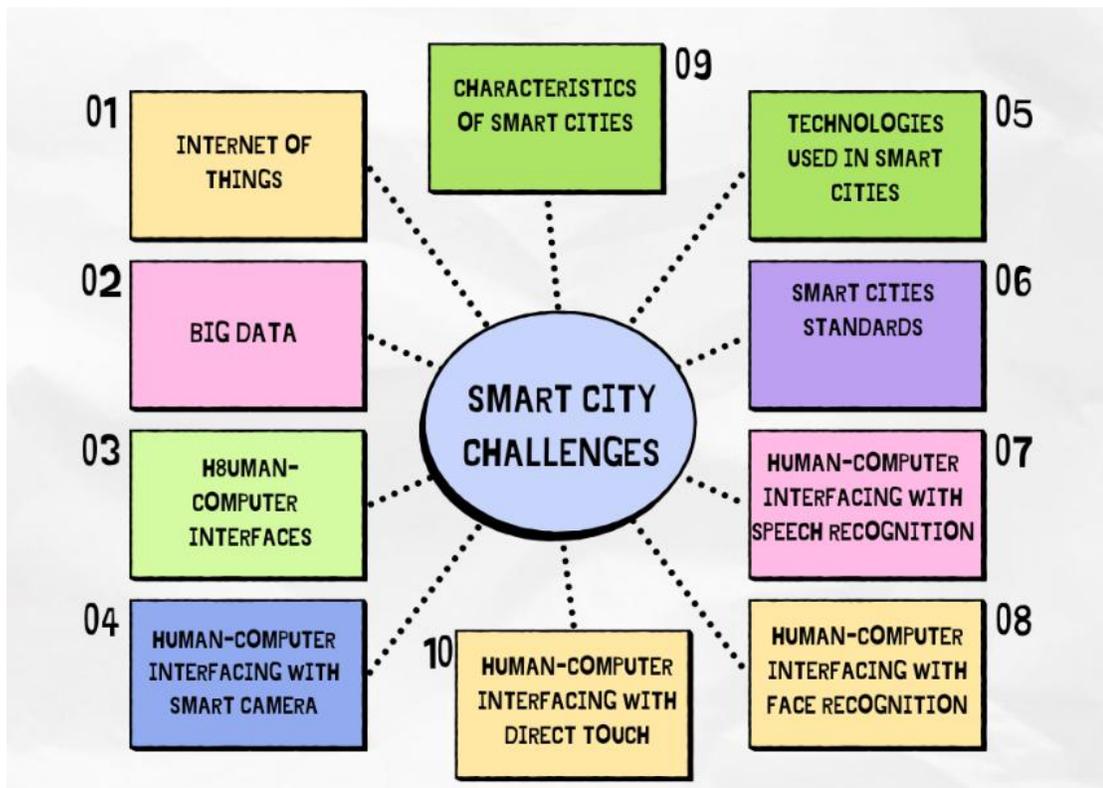

**Figure 10.** Smart Cities Challenges [172]

## 8.1 BIG DATA

Big data, IoT, and smart urban areas are firmly co-related and dependent on each other. The city information, which is set apart in reality and is created in smart urban communities, can be big data coming from sensors, messages, sites, information bases, and web-based media. The proliferation of pagers, cell phones, sensors, picture and video applications, and interpersonal organizations is delivering more than 2.5 quintillion bytes each day [173]. This bag data poses numerous challenges comprising mining, representation, catch, examination, stockpiling, sharing, and search. It requires new techniques for handling to empower further developed dynamics, vision discovery, and methodology advancement. Refined information investigation gadgets are fundamental for searching and concentrating liked plans and information from the big data of the IoT and smart urban areas. Figure 11 briefly illustrates various challenges due to this big data.

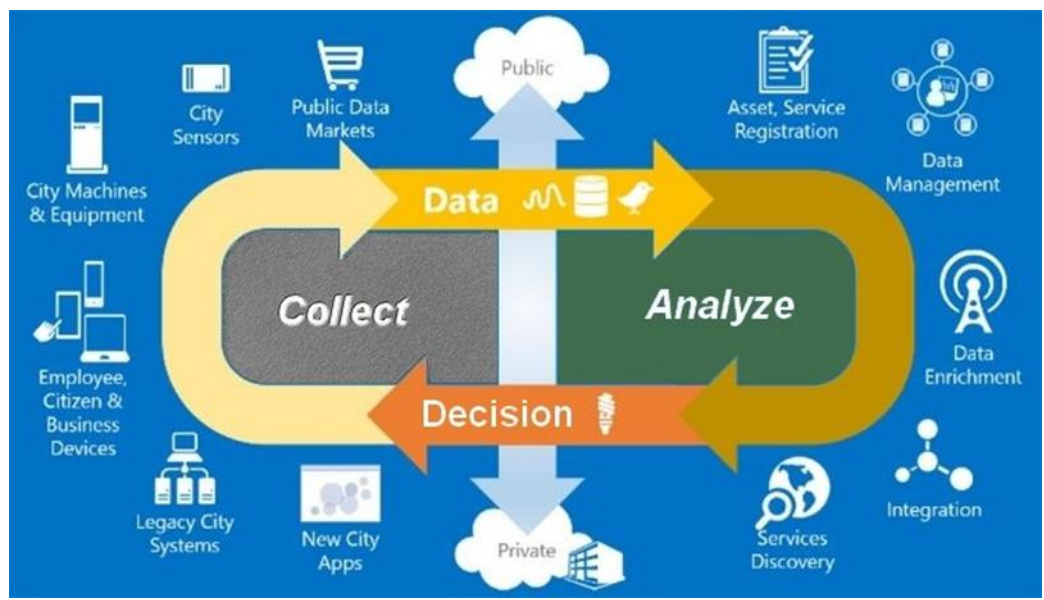

**Figure 11.** Big Data Challenges [174]

## 8.2 HUMAN-COMPUTER INTERFACES

A product, software, or program is more acceptable to its end users if its design and development are user-centered. Such a product, even with fewer features is more acceptable to users than the ones with a lot of features yet different to understand, use, and interact with, and requires training. This is where human-computer interaction (HCI) comes in, which aims to create user-centered, easy-to-use, and usable user interfaces. Smart city applications incorporate sensors, modalities that collaborate with clients, eye development filtering, hand and body motions and developments, discourse acknowledgment, bio-signals, activity acknowledgment, and numerous others. These applications must be designed from a human- and end-user perspective, achieving the envisioned dreams is almost impossible. Therefore, human-computer interfaces are much more important and thus more daunting to design and develop. A few examples are discussed in the following subsections.

### 8.2.1 Camera

The smart framework being conveyed in smart urban communities uses cameras for various purposes, including traffic management, crime control, etc. They help the client in communicating with the framework. These cameras are like visual sensors that are continuously recording data, where the challenge is to process and extract necessary information from moving captured objects, signs, and texts.

### 8.2.2 Touch

The touch screen user interfaces (UIs) not only come with smartphones, tablets, and personal computers but are also used in numerous different ventures, e.g., guides on shopping centers, transport, and rail stations that assist the client with bettering the spot. The challenge is their design, layout, and content to be put on these screens so that users get the most out of them. The UI plan, symbols, and names ought to be planned and put on UI so that they facilitate clients to go through it. There will be appropriate page routes with the right data being shown to the client instead of wrecking the screen with a heap of superfluous data. Straightforwardness is the key here when the client is going through direct touch frameworks. The client ought to have a characteristic inclination that they are controlling the framework.

### 8.2.3 Face Recognition

Smart urban areas with the help of the installed cameras can benefit from face recognition applications for improved surveillance, crime control, authentication, and authorization. However, the changes in the face of the inhabitants such as facial hair, mustache, and haircut may cause issues in face recognition. In this, the face recognition technology and algorithm must be improved to handle all such anomalies.

### 8.2.4 Speech Recognition

Speech recognition is another important aspect of smart cities and urban communities, which can be frequently seen in smartphones, intelligent cars, and smart home computerization frameworks for various activities and security purposes. However, speech recognition becomes challenged with changes in voice due to health or medical conditions that affect the sensors' ability to recognize the voice of the home or car owner. Another issue is the emphasis a user wants to put through her voice, which can be problematic for sensors and underlying applications to perceive the way it was intended.

## 8.3 STANDARDS

Worldwide smart urban communities are portrayed as gigantically different as far as their prerequisites and the parts they agree with. Nonetheless, the International Standards Organization (ISO) gives the principles that will be agreed upon to guarantee the nature of administrations, and their effectiveness just as security protection. Principles are of most extreme significance and assume an essential part in smart urban areas' advancement. They help in the formalization of the prerequisites needed for checking smart urban areas' exhibition, actually and practically. Standard is likewise used to manage the issues that run over in the improvement of smart urban communities like water issues, transportation issues, security, and environmental changes. IEEE has additionally been assuming a significant part in figuring out such guidelines that can be applied to smart urban areas including eHealth principles, brilliant matrices, and smart transportation frameworks. ISO37120 is an awesome illustration of one such norm. This standard is equipped for giving 100 markers demonstrating city execution and these furthermore contain 46 principles and 54 extra pointers. Not many of the pointers are identified with the space of financial matters, money, energy, and to wrap things up the climate. These markers are then utilized for municipal elements to gauge exhibitions, contrasting them and various urban areas' information and distinguishing proof of examples figured for improvement.

## 9 FUTURE DIRECTIONS

- Sustainable smart cities have a wider room for research and development for DL, Blockchain, FL, NLP, and LLMs. Researchers must also focus on the assimilation of semantic technologies in smart city products to enable better communication of smart devices with the consumers of the same. The use of simulated objects combined with DRL algorithms would help in making virtual representations of tangible objects so that the objects can function automatically. Some of the prospects are outlined here.
- The usability of smart devices plays a significant role, especially, when it comes to lesser technically savvy users and senior citizens, who may find using touch displays difficult to use. Integrating speech recognition and natural language understanding may improve the acceptability of these devices among the different types of users in a smart city.

- The big data produced by smart technologies, speech and face recognition systems, sensors, and other applications pose a big challenge in efficiently processing and analyzing it so that the services offered by a smart city can be better managed, such as managing traffic, transportation, energy sources, wastes, water supply, crime detection, etc.
- The role of ICT and IoT cannot be neglected in building sustainable smart cities. They have a greater potential for research and development to further improve the performance, quality, and usability of urban services while reducing costs and resource consumption. They have great potential in improving citizen-government interaction and real-time access to information.
- Rapid urbanization brings new challenges and issues that the smart cities must deal with. In the 1950s, only 31% of the world population lived in cities, which increased to 54% in 2019, and is projected to increase to 87% by 2040 [175]. Cities with a growing population and maximum use of natural resources face ecological and environmental challenges and increased public unrest. The challenges include handling overcrowding, air and water pollution, environmental degradation, contagious diseases, and crime. Steps must be taken to reduce air pollution and provide clean water, safe neighborhoods, and efficient infrastructure.
- The business, culture, transportation, entertainment, and all other areas of city life have grown inextricably linked with ICT, and the Internet has become a significant component of inhabitants' daily lives. The numerous achievements of digitizing a city's information not only provide daily comfort to the public but also construct an infrastructure and establish a concordance.
- Smart cities have greater potential to exploit technology in making the most of the available infrastructure, electricity, Internet and Wi-Fi connections, lane management, and lighting. The smarter use of technology can streamline heating, energy consumption, ventilation, and lighting by integrating solar panels into the building design and replacing traditional materials. Smart grids can be established to monitor and control energy consumption. Technology can be exploited to detect water leaks and monitor water capacity. Slow and fast lanes can be introduced for better traffic management and extended with the technology. Vehicles can be charged through charging outlets spread over the city. Green cities can be established to control air pollution, absorb $CO_2$, and emit oxygen, manage garbage using plants and smart irrigation. Wi-Fi can be used to inform users about traffic congestion, parking spaces, and other civic amenities.

- By infusing reasoning capabilities into language models, they can excel in more complex tasks, like problem-solving, decision-making, and planning. This enhances their performance on various tasks and boosts their out-of-distribution robustness. To genuinely gauge the reasoning abilities of LLMs, it is essential to consider more realistic and meaningful applications, like decision-making, legal reasoning, or scientific reasoning. When conducting research, assessing whether the specific task is meaningful and whether the proposed method can be generalized to realistic tasks and applications is vital.

## 10  CONCLUSION

This paper highlighted the significant role of Deep Learning, Federated Learning, Blockchain, NLP, and LLMs in the context of sustainable smart cities. Automating language analytics, DL, FL, Blockchain, NLP, and LLMs enhances efficiency, accuracy, and utility in various domains of smart cities. The paper explored these techniques applied across research and development, business, healthcare, media, homes, communities, and industries. Additionally, it identifies challenges such as named entity recognition, word sense disambiguation, processing metaphors, part-of-speech tagging, linguistic variations, and multilingual conversations. Despite these challenges, they remain a promising field with numerous opportunities for future research and innovation. This study serves as a foundation for further exploration of the technology applications in smart city contexts. It concludes with some future research directions for addressing challenges and enhancing these technologies' potential in smart city contexts.

## DECLARATIONS

**Funding:** This work has received no funding from any funding agency, organization, or an institution.

**Conflict of interest/Competing interests:** The authors declare that no competing interests/conflict of interests exist that may affect the content of this publication.

**Authors' contributions:** All the authors contributed equally and substantially to this manuscript.

**Data Availability:** Not applicable.

**Abbreviations**

| Terminology | Abs. | Terminology | Abs. |
|---|---|---|---|
| Artificial Intelligence | AI | Internet of Vehicles | IoV |
| Machine Learning | ML | Food Supply Chain | FSC |
| Deep Learning | DL | Long Short-Term Memory | LSTM |
| Internet of Thing | IoT | InterPlanetary File System | IPFS |
| Federated Learning | FL | Unmanned Aerial Vehicles | UAVs |
| Natural Processing Language | NLP | Blockchain Technology | BCT |
| Large Language Models | LLMs | Deep Reinforcement Learning | DRL |
| Information and Communication Technology | ICT | Quality of Service | QoS |
| Enterprise Resource Planning | ERP | Recurrent Neural Networks | RNN |
| Learning Management Systems | LMS | Support Vector Machines | SVM |
| Wireless Sensor Networks | WSN | Quality of Service | QoS |
| Vehicular Delay Tolerant Networks | VDTN | Generative Adversarial Networks | GANs |
| Interference Aware Energy Efficient Transmission Protocol | IEETP | Electronic Health Records | EHRs |
| Wireless Body Area Networks | WBAN | Semantics of Business Vocabulary and Business Rules | SBVR |
| Big Data Analytics | BDA | Unified Modeling Language | UML |
| Online Flipped Classroom Learning Method | | Chain of Thought | CoT |
| Classroom Response System | CRS | Statutory Reasoning Assessment dataset | SARA |
| Contract Understanding Atticus Dataset | CUAD | decision support systems | DSS |
| Adaptive Network Fuzzy Inference System | (ANFIS) | Research and Development | R&D |